\documentclass{article}
\usepackage{arxiv}

\usepackage{latexsym}
\usepackage{amssymb}
\usepackage{amsmath}
\usepackage{amsthm}
\usepackage{booktabs}
\usepackage{enumitem}
\usepackage{graphicx}
\usepackage{color}
\usepackage{microtype}

\usepackage{tikz}
\usepackage{dirtytalk}
\usepackage{enumitem}
\usepackage{siunitx}
\usepackage{bbm}
\usepackage{tabularx}
\usepackage{listings}
\definecolor{codegreen}{rgb}{0,0.6,0}
\definecolor{codegray}{rgb}{0.5,0.5,0.5}
\definecolor{codepurple}{rgb}{0.58,0,0.82}
\definecolor{backcolour}{rgb}{0.95,0.95,0.92}
\lstdefinestyle{mystyle}{
    backgroundcolor=\color{backcolour},   
    commentstyle=\color{codegreen},
    keywordstyle=\color{magenta},
    numberstyle=\tiny\color{codegray},
    stringstyle=\color{codepurple},
    basicstyle=\ttfamily\footnotesize,
    breakatwhitespace=true,         
    breaklines=true,                 
    captionpos=b,                    
    keepspaces=true,                 
    numbers=left,                    
    numbersep=5pt,                  
    showspaces=false,                
    showstringspaces=false,
    showtabs=true,                  
    tabsize=1,
}
\usepackage{caption}
\usepackage{multirow}
\usepackage{pdflscape}

\usepackage{tcolorbox}
\newcounter{promptboxcounter}
\renewcommand{\thepromptboxcounter}{\arabic{promptboxcounter}}
\newtcolorbox[use counter=promptboxcounter]{prompt}[2][]{floatplacement=htbp, title={\textbf{Prompt \thepromptboxcounter:} #2}, label=#1}
\usepackage{hyperref}
\usepackage{cleveref}
\usepackage{dsfont}
\usepackage{widows-and-orphans}

\usepackage[numbers]{natbib}
\newcommand{\dataset}{\textit{Quick, Draw!}}
\newcommand{\Tikz}{TikZ}
\newcommand{\concept}[1]{{\small\texttt{#1}}}
\newcommand{\suptext}{Appendix~\cite{freitas2025}}

\usepackage[acronym,hyperfirst=false,nohypertypes={acronym}]{glossaries}
\newacronym{ai}{AI}{Artificial Intelligence}
\newacronym{rdp}{RDP}{Ramer--Douglas--Peucker}
\newacronym{gpt}{GPT}{Generative Pretrained Transformer}
\newacronym{api}{API}{Application Programming Interface}
\newacronym{llm}{LLM}{Large Language Model}

\title{Relative Drawing Identification Complexity is Invariant to Modality in Vision-Language Models\thanks{This research was accepted at the 28\textsuperscript{th} European Conference on Artificial Intelligence (ECAI-2025). Paper ID: 7633.}}
\author{
 Diogo Freitas \\
 Interactive Technologies Institute\\ and NOVA LINCS \\
 Faculty of Exact Sciences and Engineering\\
 University of Madeira \\
 Portugal \\
 \texttt{diogo.freitas@staff.uma.pt} \\
 \And
 Brigt Håvardstun \\
 Department of Informatics \\
 University of Bergen \\
 Norway \\
 \texttt{brigt.havardstun@uib.no} \\
 \And
 Cèsar Ferri \\
 Valencian Research Institute for\\ Artificial Intelligence \\
 Universitat Politècnica de València \\
 Spain \\
 \texttt{cferri@dsic.upv.es} \\
 \And
 Darío Garigliotti \\
 Department of Informatics \\
 University of Bergen \\
 Norway \\
 \texttt{Dario.Garigliotti@uib.no} \\
 \And
 Jan Arne Telle \\
 Department of Informatics \\
 University of Bergen \\
 Norway \\
 \texttt{Jan.Arne.Telle@uib.no} \\
 \And
 José Hernández-Orallo \\
 Leverhulme Centre for the Future\\ of Intelligence and\\
 Valencian Research Institute for\\Artificial Intelligence \\
 Spain \\
 \texttt{jorallo@upv.es} \\
}

\begin{document}

\maketitle

\begin{abstract}
Large language models have become multimodal, and many of them are said to integrate their modalities using common representations. If this were true, a drawing of a car as an image, for instance, should map to a similar area in the latent space as a textual description of the strokes that form the drawing. To explore this in a black-box access regime to these models, we propose the use of machine teaching, a theory that studies the minimal set of examples a teacher needs to choose so that the learner captures the concept. In this paper, we evaluate the complexity of teaching vision-language models a subset of objects in the Quick, Draw! dataset using two presentations: raw images as bitmaps and trace coordinates in TikZ format. The results indicate that image-based representations generally require fewer segments and achieve higher accuracy than coordinate-based representations. But, surprisingly, the teaching size usually ranks concepts similarly across both modalities, even when controlling for (a human proxy of) concept priors, suggesting that the simplicity of concepts may be an inherent property that transcends modality representations.
\end{abstract}

\section{Introduction}

As children, when we transform images of the world into drawings and other simplified sketches, we have the intuition that some objects are simpler than others~\citep{jane1984,long2018}. For instance, six segments are enough to represent a \concept{house} that everybody can recognize, while a bit more is necessary to represent a \concept{cat}. This intuition is epitomized by some guessing games where one person picks a concept from a card deck and has to draw something quickly for their team to identify the concept. We can easily describe and recognize some very simple visual concepts, such as letters, with verbalized descriptions. For instance, the letter \concept{T} is a horizontal segment on top of a vertical segment. However,  humans struggle to describe more complex shapes with verbal descriptions~\citep{sun2022} or objects, such as a \concept{cat}, using a series of segments.

\begin{table}[htbp]
    \centering
    \caption{The simplest drawings (applying RDP algorithm on an original drawing) identified for the concept \concept{cat}. }
    \label{tab:ts-images-intro}
    \small  
    \resizebox{\columnwidth}{!}{  
    \begin{tabular}{l c c c c}
        \toprule
        Model & Original (images) & Simplified (images) & Original (coordinates) & Simplified (coordinates) \\
        \midrule

        Claude &
        \includegraphics[width=0.12\linewidth]{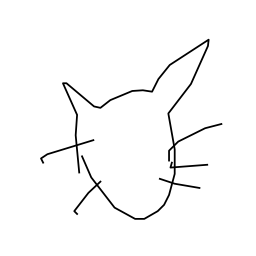} &
        \includegraphics[width=0.12\linewidth]{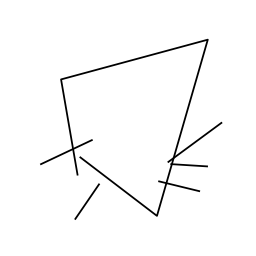} & 
        \includegraphics[width=0.12\linewidth]{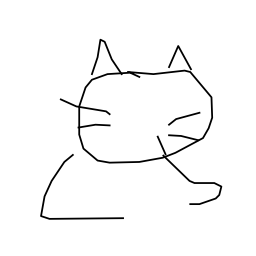} & 
        \includegraphics[width=0.12\linewidth]{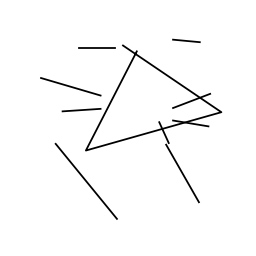} \\

        Gemini &
        \includegraphics[width=0.12\linewidth]{images/drawings/4983375733456896_e_2.png} &
        \includegraphics[width=0.12\linewidth]{images/drawings/4983375733456896_e_48.png} & 
        \includegraphics[width=0.12\linewidth]{images/drawings/5872479413207040_e_2.png} & 
        \includegraphics[width=0.12\linewidth]{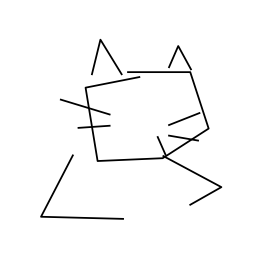} \\

        GPT-4 Turbo &
        \includegraphics[width=0.12\linewidth]{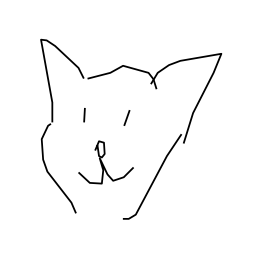} &
        \includegraphics[width=0.12\linewidth]{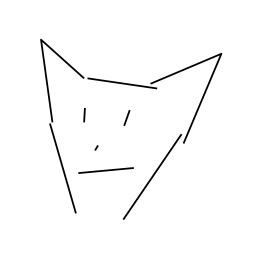} & 
        \includegraphics[width=0.12\linewidth]{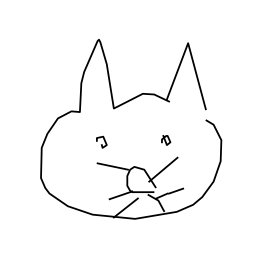} & 
        \includegraphics[width=0.12\linewidth]{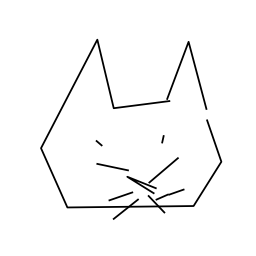} \\

        GPT-4o &
        \includegraphics[width=0.12\linewidth]{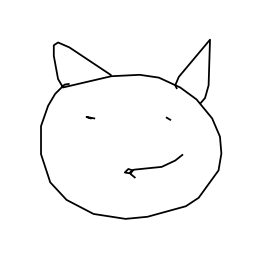} &
        \includegraphics[width=0.12\linewidth]{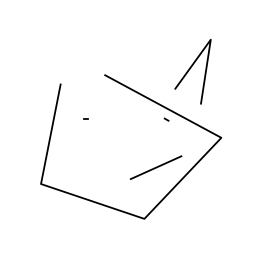} & 
        \includegraphics[width=0.12\linewidth]{images/drawings/5362546087297024_e_2.png} & 
        \includegraphics[width=0.12\linewidth]{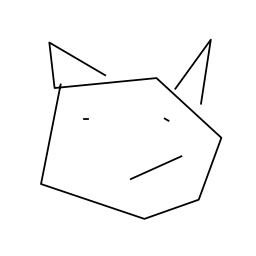} \\

        Llama &
        \includegraphics[width=0.12\linewidth]{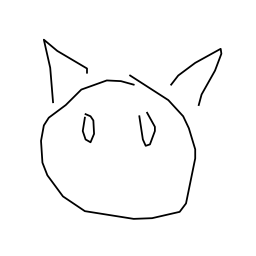} &
        \includegraphics[width=0.12\linewidth]{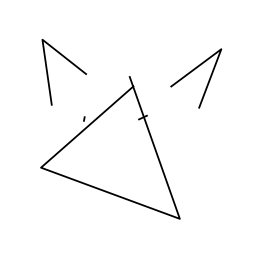} & 
        \includegraphics[width=0.12\linewidth]{images/drawings/4983375733456896_e_2.png} & 
        \includegraphics[width=0.12\linewidth]{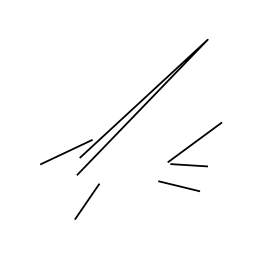} \\

        Pixtral &
        \includegraphics[width=0.12\linewidth]{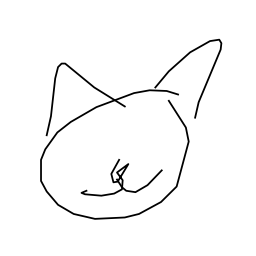} &
        \includegraphics[width=0.12\linewidth]{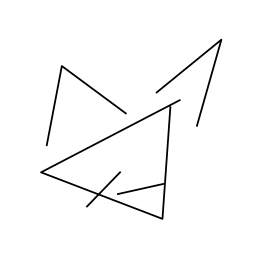} & 
        \includegraphics[width=0.12\linewidth]{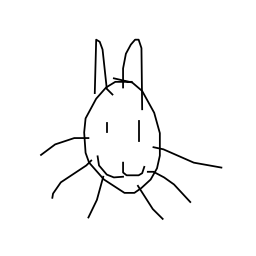} & 
        \includegraphics[width=0.12\linewidth]{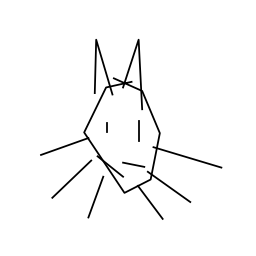} \\

        \bottomrule
    \end{tabular}
    }  
\end{table}

 \Glspl{llm} can identify objects from a textual representation of their coordinates~\citep{bubeck2023}. Thus, we aim to discover whether this understanding maps to similar capabilities for the multimodal versions of these models. Also, we do not know whether this is independent of the modality. We ask two research questions:
\begin{itemize}[leftmargin=*,noitemsep,topsep=0pt]
	\item Q1 (\textit{Absolute Invariance}): If we randomly sample a concept from a concept class, $c \in C$, would it take the same number of segments to identify it if represented as a bitmap drawing as if represented as a set of coordinates? 
	\item Q2 (\textit{Relative Invariance}): If we randomly sample two concepts from a concept class, $c_1, c_2 \in C$, 
    and $c_1$ requires fewer segments than $c_2$ when represented as a bitmap, will this order prevail when expressed as coordinates? 
\end{itemize}

\begin{figure*}[htbp]
	$\:\:\:\:\:\:\:\:$\footnotesize{Bitmap:}$\:\:\:\:\:\:\:\:\:$
	\hspace{0.15cm}
	\begin{minipage}{0.35\textwidth}
		\centering
		\includegraphics[width=0.30\linewidth]{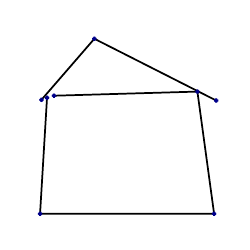} 
	\end{minipage}
	\hspace{0.15cm}
	$<$
	\hspace{0.15cm} 
	\begin{minipage}{0.35\textwidth}
		\centering
        \hspace{3.5cm}
		\includegraphics[width=0.30\linewidth]{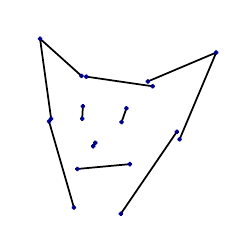} 
	\end{minipage}
	\vspace{0.5cm}
	\newline
	\noindent 
	$\:\:\:\:\:\:\:\:$\footnotesize{Coordinates:}$\:\:\:\:\:\:\:\:\:$
	\hspace{0.15cm}
	\begin{minipage}{0.35\textwidth}
		\scriptsize{\texttt{\textbackslash draw (10, 169) -- (0, 0) -- (250, 0) -- (226, 178) -- (20, 172); \\
			\textbackslash draw (2, 166) -- (78, 255) -- (253, 165);
		}} \\
		\begin{center} \footnotesize{\concept{House}} \end{center}
	\end{minipage}
	\hspace{0.15cm}
	$<$
	\hspace{2cm}
	\begin{minipage}{0.35\textwidth}
		\scriptsize{\texttt{\textbackslash draw (49, 9) -- (13, 134);\\
			\textbackslash draw (67, 199) -- (163, 185);\\
		}}
		\dots \\
		\begin{center}  \footnotesize{\concept{Cat}} \end{center}
	\end{minipage}
    \vspace{0.1cm}
	\caption{In this paper, we address two research questions. First, Q1 (absolute invariance): When using a vision-language model, are bitmaps (top) equally efficient representations for drawings as coordinates (bottom)? The second question is Q2 (relative invariance): Is the order (left vs. right) of simplicity preserved across modalities?}
	\label{fig:firstpage}
\end{figure*}

Question Q1 refers to whether a concept represented as a bitmap drawing is easier or harder to recognize than the same concept as coordinates in text, while question Q2 is about the relative ranking. For instance, consider that $c_1$ is a \concept{house} and $c_2$ is a \concept{cat}. In Figure~\ref{fig:firstpage}, if a \concept{house} is easier than a \concept{cat} when using the bitmap of the drawing (top of the figure), is it also easier when represented as segment coordinates (bottom of the figure)? This is the \emph{relative invariance}.
%
%
Note that we are not comparing with photographic images of the object since other features would come into play, such as a striped texture to distinguish a \concept{tiger} from other felines. 
Such distinctions are particularly evident in machine vision systems~\citep{geirhos2023}. 

However, how can one determine the notion of simplicity of a concept from its drawings? The idea we pursue in this paper is based on the field of machine teaching~\citep{zhu2018}, and in particular, the notion of teaching minimality. A concept is as simple as a teacher can communicate the concept to a learner with as little information as possible. This captures our intuition that a \concept{house} needs six segments while a \concept{cat} needs  more segments. Given a concept, the teacher 
has to find the simplest drawing in terms of the number of straight-line segments---the teaching size---that enables the learner to consistently recognize the concept. 
We use two different types of language representations (bitmaps of the drawing and coordinates in \Tikz{} code) to present the concepts to the learner. Multiple models, including \Gls{gpt}-4~\citep{achiam2023}, Llama~\citep{grattafiori2024}, Gemini~\citep{gemini2024}, Pixtral, and Claude, are employed as the \say{learners}. The resulting collection of the simplest images identified, across all concepts, all modalities, and all models, is intriguingly diverse. As a preview of our findings, see Table~\ref{tab:ts-images-intro}, showing the simplest identified images for the concept \concept{cat}. 

It is also important to note that priors play a role in machine teaching. When in doubt, the learner will more likely associate the evidence with the most common concept (e.g., a \concept{house} is more common than an \concept{envelope}). Accordingly, a Bayesian prior will be used to disentangle this effect when looking at the concept simplicity rankings.

The contributions of this paper are: 
\begin{itemize}[leftmargin=*,noitemsep,topsep=0pt]
	\item A novel machine teaching framework for evaluating the complexity of concepts, which can be applied to drawings in coordinate- and image-based modalities.
	\item Use of the teaching size specifically to evaluate how simply and effectively the concept can be taught across both modalities.
	\item A comparison of both modalities across multiple models, including \Gls{gpt}-4, Llama, Gemini, Pixtral, and Claude, 
    according to the number of concepts identified, accuracy, frequency of errors, and teaching size.
	\item A way to disentangle the effect of the learner's prior knowledge in the concept identification task. 
\end{itemize}

These contributions are generic and can be applied to other problems and modalities. In our particular case, we show that bitmaps are more efficient than coordinates, but surprisingly, the order of complexity between the concepts is preserved to some extent. This suggests that either the representations of both modalities are tightly connected in the latent space of the model, or the simplicity of concepts is an inherent property that transcends modalities.

\section{Related Work}

\textbf{Drawing (or Sketches) Recognition:}
\citet{eitz2012} 
provided a dataset of human drawings, including \num{250} concepts and \num{20000} drawings. 
They introduced a support vector machine model to recognize these drawings and observed that humans outperformed its performance. Since then, AI models have been closer or even achieved higher accuracy than that of human classification for drawing recognition (e.g.,~\citealt{schneider2014,yu2015,zhang2020,yang2024}). Using the~\dataset{} dataset, \citet{ha2017} proposed \texttt{sketch-rnn}, a 
model designed to create drawings of common objects that resemble those drawn by humans. A similar version of this model has also shown capabilities in drawing recognition~\cite{payal2017}. Other neural approaches studied for this task include convolutional neural networks~\cite{kabakus2020}, and graph neural networks applied over drawings represented as graphs~\cite{xu2022}.

\textbf{Drawing Capacities of LLMs:}
%
\citet{sharma2024} assess the visual abilities of different language models. 
They conduct experiments that prompt the models to create code that draws images based on text descriptions and improve image generation code iteratively through text feedback. 
They show that: (a) \Gls{llm}s 
possess limited ability to recognize concepts represented in code, and (b) these models sometimes fail to recognize concepts that they can accurately draw.  
Note that the authors addressed the problem as a multi-class classification problem. Moreover, the online interface for collecting human drawings limits components to basic shapes like ellipses, possibly restricting participants' ability to create complex drawings. In their initial experiments with \Gls{gpt}-4, \citet{bubeck2023} present an example of drawing generation, showcasing text-to-image capabilities using \Tikz{}. They show tasks such as \Gls{gpt}-4 drawing a unicorn and constructing \Tikz{} code through a multi-step prompt process. In another study,~\citet{pourreza2023} introduce the \textit{Painter}, a modified \Gls{llm} that creates drawings using virtual brush strokes based on user-provided text descriptions. 
Additionally,~\citet{cai2023} evaluated \Gls{gpt}-4's ability to understand visual data in SVG format across various visual tasks, including image classification, visual reasoning, and image generation. \citet{vinker2024} propose \textit{SketchAgent}, showing that while LLMs iteratively generate sketches, they struggle with spatial reasoning.

\textbf{Machine Teaching:}
%
Machine teaching is a research area that focuses on identifying the optimal set of examples that allow a learner (e.g., a human or a machine) to identify a given concept~\cite{zhu2018}. 
To illustrate the underlying idea of machine teaching, assume the teacher wants the learner to identify the concept of prime numbers. To achieve this, the teacher uses the set $S_1 = \{2,3,5,7,11,13\}$ and succeeds. However, would it not be enough for the learner just to see the smaller set $S_2 = \{19,23\}$? Of course, that depends on the learner. In general, optimal teaching will depend on the model the teacher has of the learner. Machine teaching presents an alternative framework to machine learning (where examples are not chosen but sampled from a distribution) to answer the question of whether some concepts are inherently more complex than others. The connections between machine teaching and computational learning theory are strong; see, e.g., the works by~\citet{doliwa2014} or~\citet{moran2016}, with machine teaching putting the emphasis on the minimal evidence that distinguishes the concept from all the rest. To determine how easy it is to teach a concept, the teaching dimension~\cite{zhu2018}---the minimum number of examples the learner needs to identify a concept---was traditionally used. 
~\citet{telle2019} introduced a new metric named teaching size. This metric puts the focus on the sum of the sizes of the examples needed to identify a concept, rather than only the number of examples. 

\section{Methods}

The drawings used in this work come from the \dataset{} dataset~\cite{jongejan2016,ha2017}, which includes over \num{50} million drawings of \num{345} concepts. Collected by Google Creative Lab via an interactive game, participants had \num{20}~seconds to draw a concept while a neural network attempted real-time recognition. The dataset is the largest collection of doodles in the world, with contributions from more than \num{15}~million participants.

Each drawing in the Simplified Drawing files that we use is stored as vectors of distinct pen strokes, i.e., distinct continuous movements of the pen without lifting. Each stroke $s_i$ is represented by a sequence of $(x, y)$ coordinates $\{(x_{i1}, y_{i1}), (x_{i2}, y_{i2}), \ldots, (x_{in}, y_{in})\}$. Note that each pair of consecutive points in a stroke creates a segment. Additionally, for each drawing, a binary flag $r$ indicates whether the game's neural network correctly recognized the concept.

The following sections cover concept selection, corresponding drawings, learners, the machine teaching setting, and the drawing selection conducted before testing the framework.

\subsection{Teaching Size}

Let $D$ denote an infinite space of possible drawings (and their simplifications, as will be explained later), and let $C$ be a set of concepts. We use $D_c$ to denote all the drawings of a concept $c \in C$. For any given concept $c \in C$, the objective is to identify the simplest drawing $S \in D_c$ (represented as $S^m$ with modality $m$ being either bitmap or coordinates) such that a learner $L$ successfully learns $c$ with a probability of at least $\rho$ over $N$ independent trials (i.e., recognition consistency). The \emph{teaching size (TS)} of $c$ for the modality $m$ can then be defined as follows:
\begin{equation}\label{eq:ts}
    \begin{aligned}
        \text{TS}_{\rho,N,m}(c) = \min_{S \in D_c} |S^m| \: 
        \text{s.t.} \: \sum^{N}_{1} \mathds{1} \left[ 
        L(S^m) = c \right] \geq \rho \cdot N
    \end{aligned},
\end{equation}
where $\mathds{1}[\cdot]$ is the indicator function, which equals 1 if the learner $L$ correctly identifies concept $c$ from the drawing $S^m$, and 0 otherwise.

We argue that a good metric for assessing the simplicity of a given drawing $d$ can be based on the number of segments it contains. This is represented by $|S^m|$ in the above equation. This metric is intrinsic to the drawing itself, thereby avoiding dependencies on the length or verbosity of the instructions used to generate it, such as in a descriptive language like \Tikz.

We also note here that while our implementation of teaching size is grounded in segment count for drawings, the framework itself is more general. Teaching size, as a proxy for descriptive complexity, can be adapted to other domains using modality-appropriate metrics.

\subsection{Concepts}

In our work, if the expected concept is \concept{car} and the identified concept is \concept{police car}, the identification is still considered correct because \concept{police car} is a specific type of \concept{car}, i.e., it is a semantically related prediction. This approach is similar to the one followed by Lamb et al. (\citeyear{lamb2020}). 
This means that if a specific sub-concept, or \emph{hyponym}, is identified, it should still be seen as a correct identification as long as it falls under the more general expected concept. 
For a concept $c$, such as \concept{car}, we consider a set of hyponyms $h(c)$ that corresponds to a set of concepts with a more specific meaning than $c$, e.g., \concept{police car} belongs to $h($\concept{car}$)$.
For this study, we want a set of concepts that ensures that in the set of their hyponyms, there is no overlap, i.e., for any two concepts $c_i,c_j$, we have $h(c_i) \cap h(c_j) = \emptyset$. This rules out certain pairs of concepts available in the \dataset, like \concept{van} and \concept{car}, and it enhances the clarity and robustness of the study. 
We thus select the following subset of \num{20} concepts from the \num{345} concepts available in \dataset{}, with no overlap among their hyponyms: \concept{apple}, \concept{banana}, \concept{car}, \concept{cat}, \concept{computer}, \concept{cup}, \concept{door}, \concept{envelope}, \concept{fish}, \concept{grass}, \concept{hockey puck}, \concept{house}, \concept{key}, \concept{radio}, \concept{string bean}, \concept{sun}, \concept{sword}, \concept{television}, \concept{The Great Wall of China} and \concept{tree}.

In Table~3 in the \suptext, we list each concept from the dataset and the accepted hyponyms that are considered correct. 
This correspondence is established by human inspection and after the execution of the drawing selection phase (cf. Sect.~\ref{method:preframework}) and the machine teaching framework experiments, with the results then analyzed based on these mappings.

\subsection{Drawings}


After choosing the concepts to study, we only include drawings that the game's neural network correctly identified (i.e., $r=1$) in our research. For every concept, approximately \num{50} drawings are selected by a proportional random stratified sampling method~\cite{taherdoos2016}, which groups drawings into bins based on their number of segments. (This number is approximate, as there may be rounding errors when calculating the number of samples for each bin according to its proportion.) The bin width was obtained using the minimum bin width between the Sturges's rule and the Freedman--Diaconis Estimator, ensuring that drawings of any concept are represented in a way that reflects the distribution of stroke counts for all correctly identified drawings of that concept in the dataset.

\begin{figure*}[htbp]
    \centering
    \begin{tikzpicture}
        \node (img1) at (-2, 2) {
            \begin{tabular}{c}
                \includegraphics[width=0.1\textwidth]{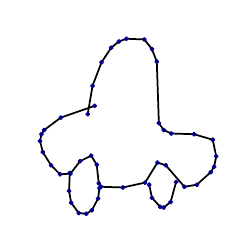} \\
                \text{\small $\epsilon=2$; 48 segments}
            \end{tabular}
        };
        \node (img2) at (1.7, 2) {
            \begin{tabular}{c}
                \includegraphics[width=0.1\textwidth]{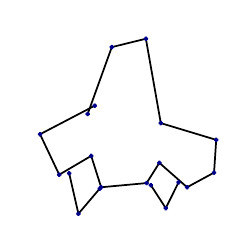} \\
                \text{\small $\epsilon=13$; 17 segments}
            \end{tabular}
        };
        \node (img3) at (5.4, 2) {
            \begin{tabular}{c}
                \includegraphics[width=0.1\textwidth]{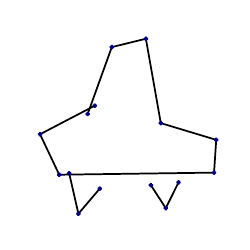} \\
                \text{\small $\epsilon=27$; 12 segments}
            \end{tabular}
        };
        \node (img4) at (9.1, 2) {
            \begin{tabular}{c}
                \includegraphics[width=0.1\textwidth]{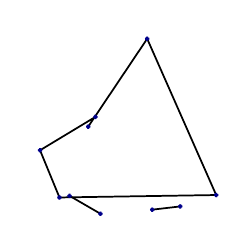} \\
                \text{\small $\epsilon=46$; 7 segments}
            \end{tabular}
        };
        \draw[->, thick] (img1) -- node[midway, above] {\Gls{rdp}} (img2);
        \draw[->, thick] (img2) -- node[midway, above] {\Gls{rdp}} (img3);
        \draw[->, thick] (img3) -- node[midway, above] {\Gls{rdp}} (img4);
    \end{tikzpicture}
    \caption{Example of a drawing simplification for the concept \concept{car} using the \Gls{rdp} algorithm. As the value of $\epsilon$ increases, the drawings become progressively simpler.
    }
    \label{fig:drawing-simplification}
\end{figure*}
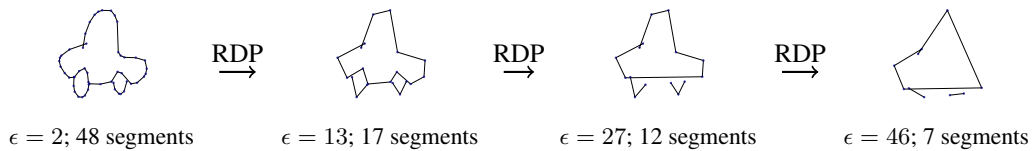

To simplify the drawings in our study, we employ the \Gls{rdp} algorithm~\cite{ramer1972,douglas1973} on each stroke $s$ of a given drawing $d$. \Gls{rdp} reduces the number of segments in each stroke while preserving its overall shape. Specifically, given a stroke $s$ with a sequence of points $\{(x_{1}, y_{1}), (x_{2}, y_{2}), \ldots, (x_{n}, y_{n})\}$, the \Gls{rdp} algorithm iteratively selects the most distant point $(x_{d}, y_{d})$ from the line segment connecting the first and last points of the stroke. If this distance is below a predefined threshold $\epsilon$, then this stroke is simplified to a single segment $\{(x_{1}, y_{1}), (x_{n}, y_{n})\}$ on the first and last points.
However, if the distance to $(x_{d}, y_{d})$ exceeds $\epsilon$, the algorithm keeps this point and recursively processes the two sequences of points formed by $\{(x_{1}, y_{1}), \ldots, (x_{d}, y_{d})\}$ and $\{(x_{d}, y_{d}), \ldots, (x_{n}, y_{n})\}$. 
This ensures that the essential characteristics of the stroke, up to distance $\epsilon$, are preserved. This process continues until all points in the stroke fall within the threshold, resulting in a simplified representation of the stroke with fewer segments.
By incrementing the threshold parameter, from an initial value of $\epsilon = 2$ \footnote{The strokes stored in the Simplified Drawing files of \dataset{} have already been simplified by the \Gls{rdp} algorithm using $\epsilon=2$, so this initial value did not simplify any drawing further.}, until each stroke is reduced to one segment, we generate simplified versions of each original drawing associated with a given concept $c$, resulting in new drawings $\{d\}_{\epsilon} \subseteq D_c$. Figure~\ref{fig:drawing-simplification} illustrates a drawing simplification.

We note here that image and coordinate representations are generated differently, but both encode the same visual information. While not equivalent in all respects, the coordinates in \Tikz{} are a form of structured data that, by reflecting a sequence of drawing actions, yield the same shape as the image once rendered. 

\subsection{Learners \texorpdfstring{$(L)$}{(L)}}

We utilize multiple \Glspl{llm}, including two \Gls{gpt}-4 models (\texttt{gpt-4-turbo} and \texttt{gpt-4o}) from OpenAI, Llama (\texttt{Llama-3.2-90b-vision-instruct}) from Meta, Gemini (\texttt{gemini-pro-1.5}) from Google DeepMind, Pixtral (\texttt{pixtral-large-latest}) from Mistral, and Claude (\texttt{claude-3-5-sonnet}) from Anthropic. These models are capable of processing visual and language inputs to produce text outputs. To conduct the experiments of this work, all models were accessed via their respective \Glspl{api}. Additionally, we set the temperature parameter $T$ to 1 for the experiments carried out within the machine teaching framework, and we set $T=0$ for the drawing selection phase. $T \in [0..2]$ controls the behavior of the models' outputs: the lower $T$ is, the more deterministic (predictable) results it leads to~\cite{openai2024}. Thus, by setting $T=0$ in the drawing selection phase, our goal is to obtain deterministic and predictable results, which are essential for creating a consistent baseline of drawings where the concepts were correctly identified. On the other hand, setting $T=1$ in the experiments of the machine teaching framework is intended to introduce a controlled level of variability. 

We consider two different representations for each concept: a visual representation and a text-based representation. Accordingly, we develop and test two prompt templates, one for each modality. For the vision-based modality, the drawings are presented as images generated from the sequence of coordinates (cf. Prompt~1 in the \suptext). For the text-based modality, the pen stroke vectors are coded using the \Tikz{} language (cf. Prompt~2 in the \suptext). Both prompts ask for an open-ended answer (not multiple choice), allowing the learners to consider a wide range of possible concepts when identifying a given concept, including any that is not in our \num{20}-concept set.

Data contamination occurs when language models are tested and evaluated using information from their training data, such as drawings already seen during training~\cite{ravaut2024}. However, in this study, the drawings are consistently simplified using the \Gls{rdp} algorithm. This algorithm alters the coordinate information, thereby modifying the \Tikz{} code and the visual representation. Consequently, we argue that these modified drawings are not part of the training set used to train the learners. Therefore, contamination tests are not required for this experiment.

It is important to note that although the models are not trained during our experiments, we refer to them as \say{learners}, since this is aligned with the standardized terminology of machine teaching.

\subsection{Concept Priors}

As we argue in the introduction, some concepts, such as a \concept{house}, are more common than others, such as an \concept{envelope}. This sets a strong prior bias, especially in cases of doubt. 
For each of the \num{20}~concepts, we use the 2022 English corpus of Google Books Ngram
~\cite{googlengram2010}, 
providing the prior of a given concept as a normalized number between \num{0} and \num{1}, representing the relative frequency of the concept. 
The rationale for using word frequency from Google Books Ngram as a proxy for human priors lies in the historical and cultural representativeness of a corpus. The assumption underlying our approach is that the frequency of specific words and phrases in written text correlates with their prominence in human thoughts, discussions, and collective knowledge at particular times~\cite{tanaka2011}. Given that LLMs 
are trained on large text corpora that include books, articles, and other written materials, it is reasonable to assume that the Google Books Ngram priors closely align with the priors embedded in LLMs. 

The priors were obtained in a case-insensitive manner. Each concept is treated exclusively as a noun to prevent confusion with its verb form (i.e., \concept{fish} is interpreted as the animal and not the fishing activity).

\subsection{Drawing Selection Phase}
\label{method:preframework}

Before applying the machine teaching framework, we first conduct a drawing selection phase. This process identifies which drawings are reliably recognized by each model across modalities. These filtered examples form the basis for estimating teaching size. Hence, our minimization of Eq.~\ref{eq:ts} is sufficiently accurate. 

As already mentioned, the drawings are simplified using the \Gls{rdp} algorithm, starting with a threshold of $\epsilon = 2$ on the raw drawings and continuing until each stroke in the drawing consists of a single segment. For each $\epsilon$, the learner is prompted using Prompt~1 for visual-based identification and Prompt~2 for text-based identification (cf. \suptext). Then, based on the completions from the learner, we obtain, by human inspection, the correspondence (between concepts and their respective accepted hyponyms) described in Table~3 in the \suptext, and we analyze the results based on those mappings. The accuracy and frequency of mistakes for each concept are  obtained from the drawing selection phase.

In total, for the drawing selection phase, we run tests on each learner separately, generating a total of $21,896$ prompts—half ($10,948$) for coordinates and half for images. These prompts were checked by human visual inspection, producing Table~3 (\suptext). We then use the drawings that are correctly identified to test and evaluate the machine teaching framework proposed in Eq.~\ref{eq:ts}, and thus obtain, for each concept, the teaching size.

\section{Results}

\subsection{Concepts Identified}

Out of the \num{20} concepts evaluated, all were identified in the image-based modality by at least one model. However, for the coordinates representation, \concept{television}, \concept{sword}, \concept{radio}, \concept{car}, \concept{door}, \concept{hockey puck}, \concept{string bean}, and \concept{The Great Wall of China} were never recognized by any model. We hypothesize that not only the complexity but also the prior of each of these latter concepts is behind their failed identification.

The image-based modality is thus more effective than the coordinate-based modality in identifying a broader range of concepts. This observation aligns with the typical human learning patterns, where visual information is often easier to process and understand than abstract textual-numerical data.

\subsection{Accuracy}

\begin{figure*}[htbp]
	\centering
        \begin{minipage}{0.45\textwidth}
		\centering
        \includegraphics[width=.85\linewidth]{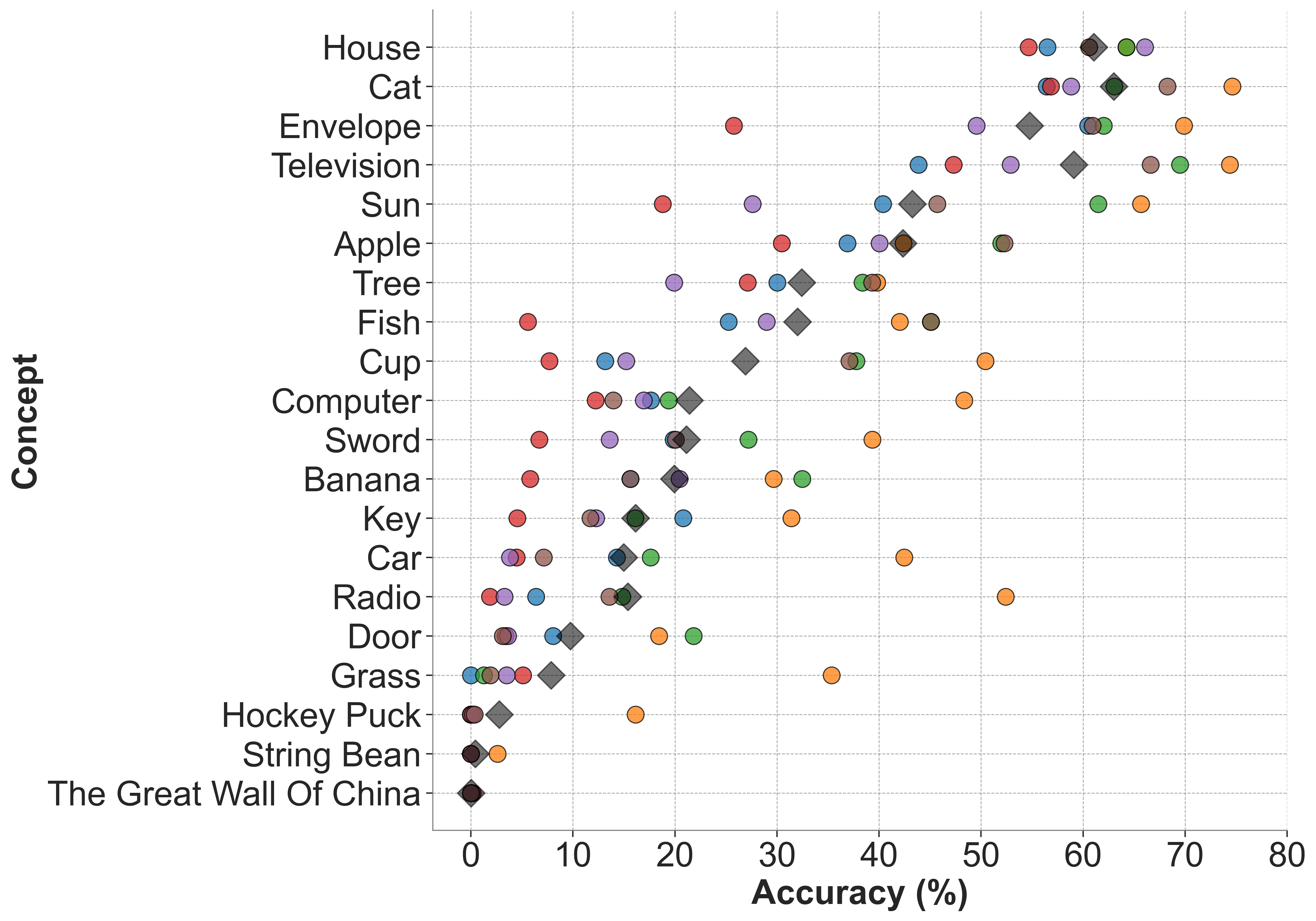} 
	\end{minipage}
    $\:\:\:\:\:\:$
	\begin{minipage}{0.45\textwidth}
		\centering
        \includegraphics[width=.85\linewidth]{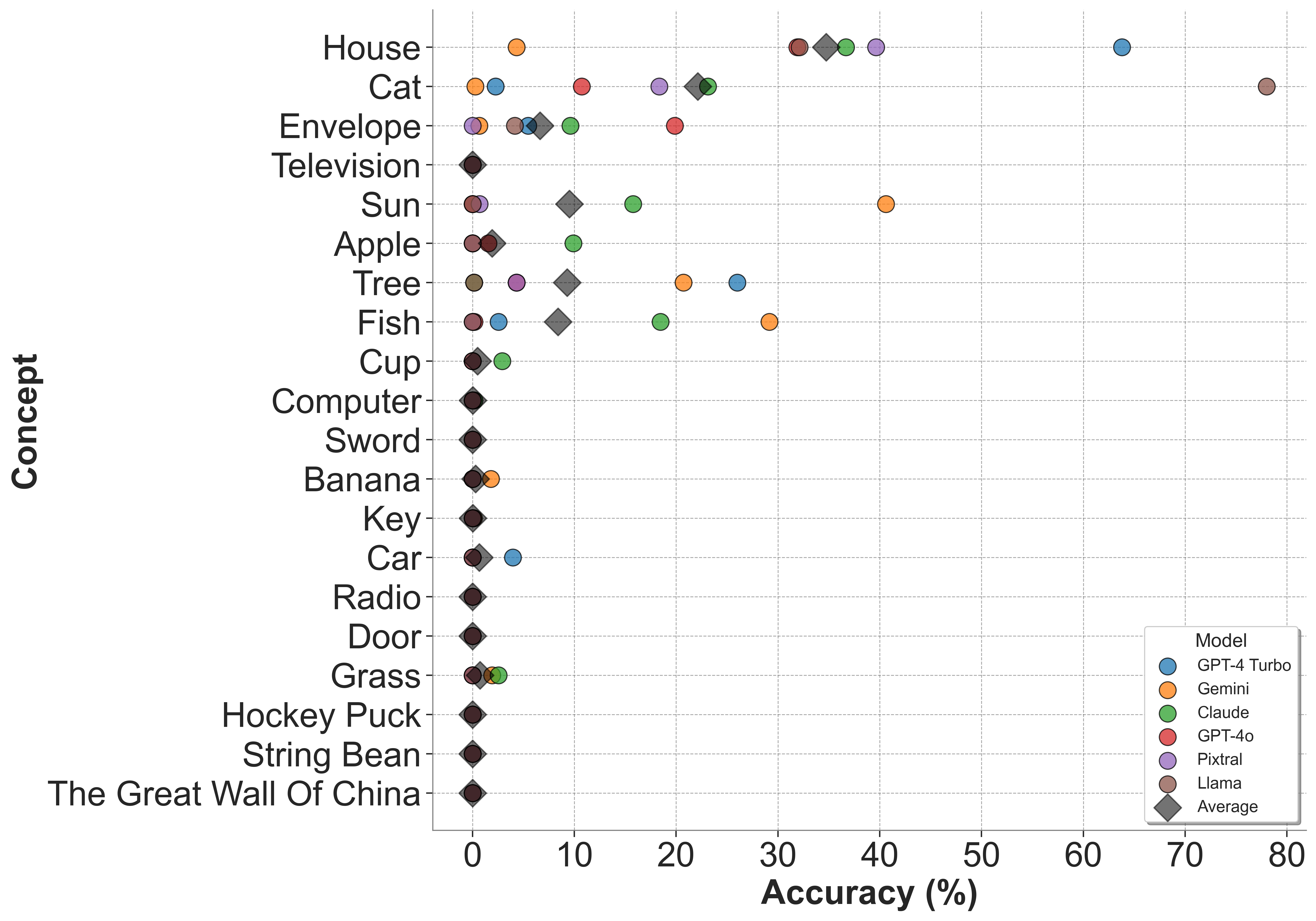}
	\end{minipage}
	\caption{Accuracy for each concept in the vision-based (images; left) and text-based (coordinates; right) modality representations. $\blacklozenge$ represents the average accuracy value for the concept.}
	\label{fig:acc-both-modalities-side-by-side}
\end{figure*}

We begin by evaluating the accuracy on each concept $c$, $\text{Accuracy}(c)$, defined here as
\begin{equation}
    \text{Accuracy}(c) = \frac{1}{N_c} \sum_{i=1}^{N_c} \mathds{1} \left[ L(S_i) = c \right],
\end{equation}
where $N_{c}$ corresponds to the total number of tests (in this case, prompts) conducted on $L$ for the concept $c$ on the drawing selection phase, with $\{S_i\}_{i=1}^{N_{c}} \subseteq D_c$.

Figure~\ref{fig:acc-both-modalities-side-by-side} depicts each concept's accuracy across the two modality representations. 
The image modality 
shows a wider range of recognition accuracy, with average performance metrics spanning up to 65\%. In contrast, the coordinate modality exhibits a much narrower range, largely confined to 0--25\% average accuracy. This discrepancy likely reflects the models' ability to leverage richer visual features in image-based representations compared to the sparse and abstract nature of coordinate-based inputs. The richer detail in images provides more cues for concept identification, while the textual coordinates impose a more constrained and abstract recognition task. 

Nevertheless, the results suggest that some concepts are fundamentally challenging to recognize, regardless of the modality. 
\concept{House} and \concept{cat} achieve relatively high accuracy across both modalities, indicating their simplicity or recognizability regardless of representation. In contrast, more complex or less visually distinct concepts, such as \concept{hockey puck} and \concept{The Great Wall of China}, show zero accuracy in the coordinate modality and only marginal performance in the image modality. 


Among the models evaluated, Gemini emerges as the best-performing model in the image modality, consistently achieving better results across a broader range of concepts. Notably, it stands apart from the other models, which appear to form a distinct cluster in terms of performance. This suggests that while certain concepts are uniformly challenging across all models, Gemini is better equipped to handle a wider range of visual representations. In the coordinate modality, no single model shows clear superiority, likely due to the shared constraints of the textual representation.

The precise accuracy of each concept, categorized by model and modality, can be found in Table~4 in the \suptext.

\begin{figure*}[htbp]
	\centering
        \begin{minipage}{0.45\textwidth}
		\centering
        \includegraphics[width=0.80\linewidth]{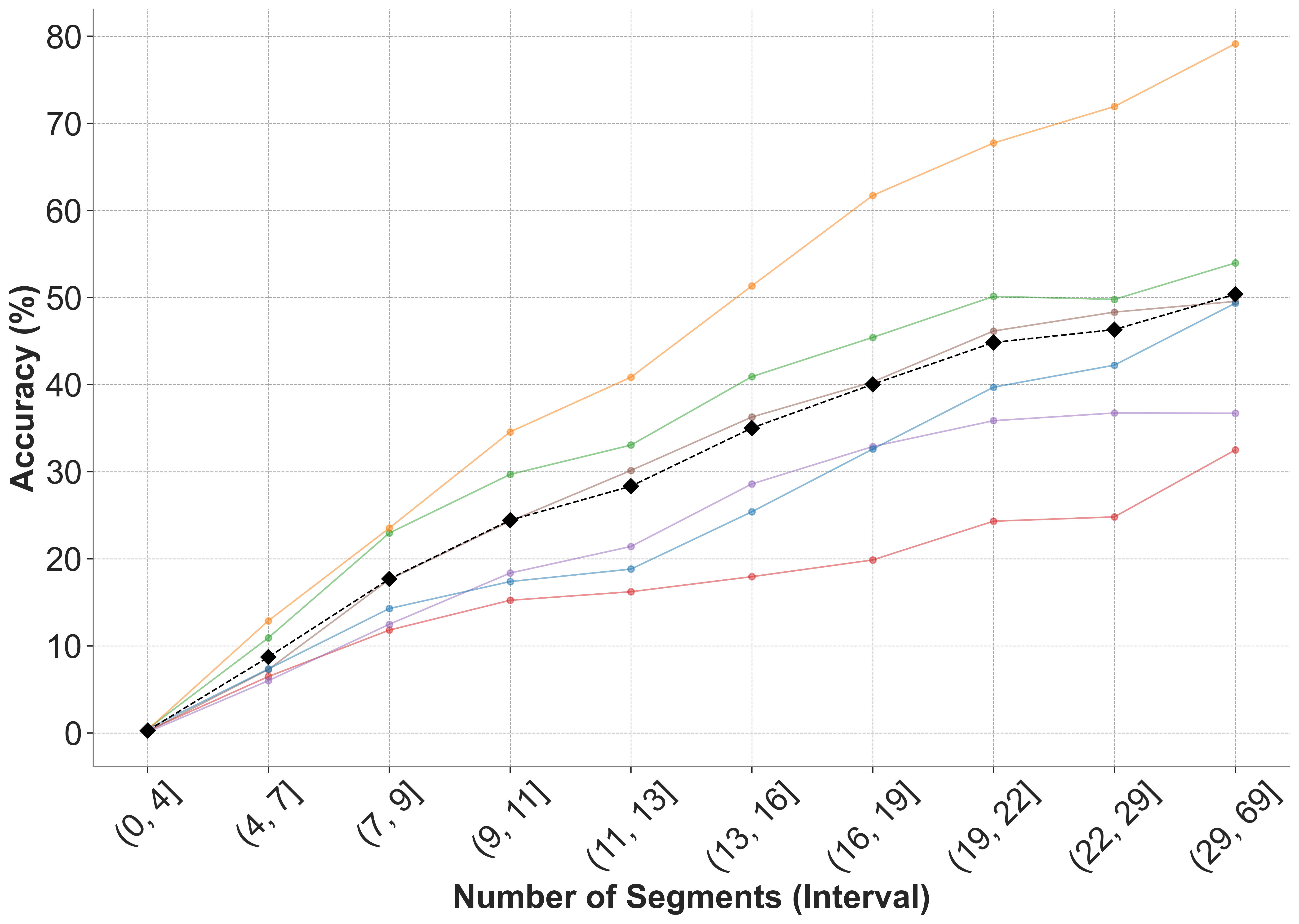}
	\end{minipage}
	$\:\:\:\:\:\:$
      \begin{minipage}{0.45\textwidth}
		\centering
        \includegraphics[width=0.80\linewidth]{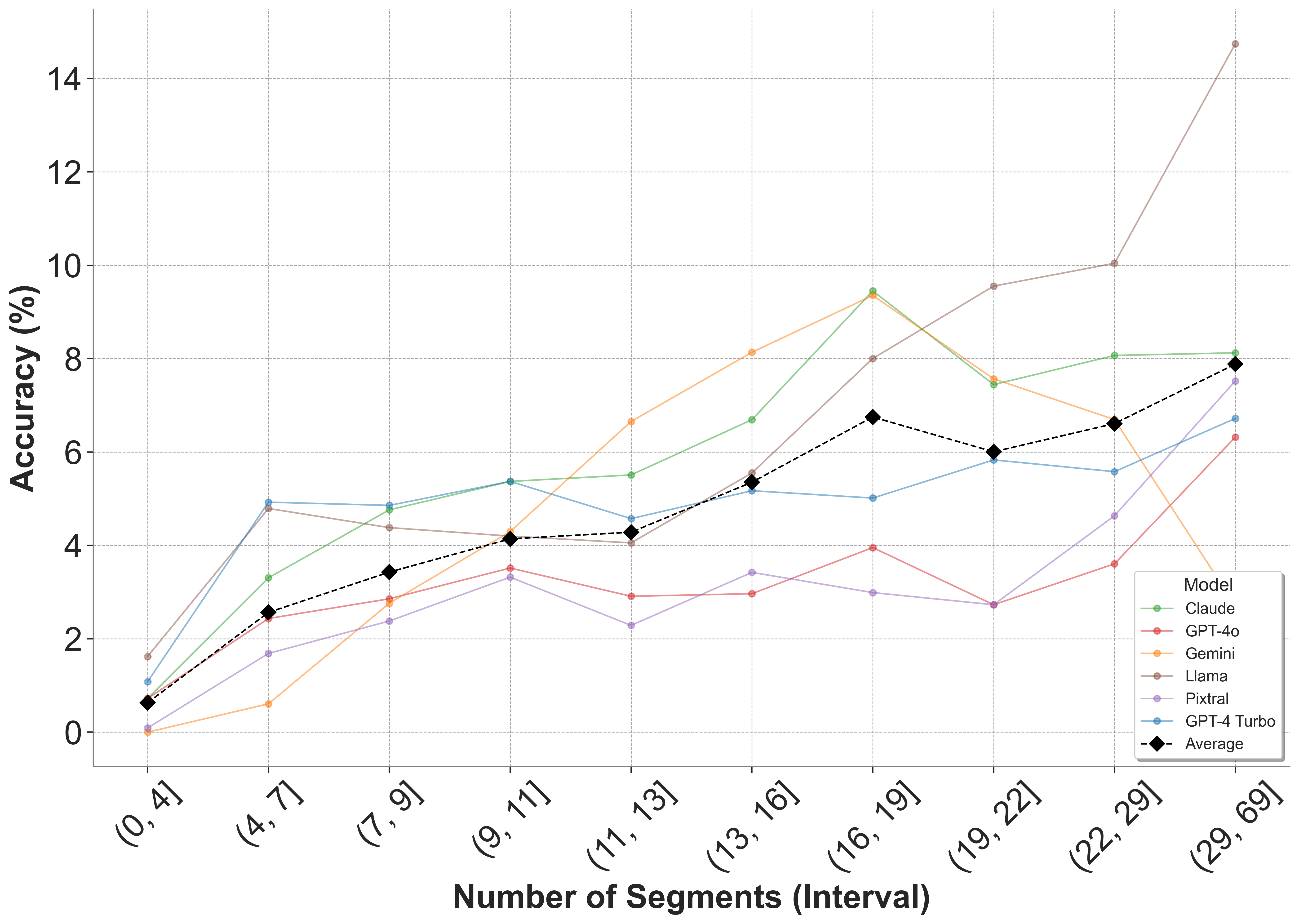} 
	\end{minipage}
	\caption{Relationship between the number of segments and accuracy for both modalities (images; left) (coordinates; right).}
	\label{fig:acc-quantile}
\end{figure*}

We also study the relationship between the number of segments (i.e., complexity) and the accuracy of concept identification for both image- and coordinate-based representations, as shown in Figure~\ref{fig:acc-quantile}. For image-based representations, there is a clear positive relationship between accuracy and the number of segments. Starting from an accuracy of around \qty{0.3}{\percent} in the $(0, 4]$ interval, the accuracy increases steadily, reaching approximately \qty{50}{\percent} in the $(29, 69]$ interval.

Conversely, for coordinate-based representations, the average accuracy remains significantly lower and follows a more modest increasing trend. Beginning at roughly \qty{1}{\percent}, it gradually rises to around \qty{8}{\percent} in the $(16, 19]$ interval before stabilizing and fluctuating slightly in the higher segment intervals. This indicates that increasing the number of segments in coordinate-based representations provides only minimal benefits 
in accuracy.

\subsection{Frequency of Mistakes}

Accuracy measures how well the learner has identified the correct concepts. However, the model can also respond with \say{I don't know} answers (or something that is not a concept) or by identifying a different concept that is incorrect. We focus on the latter case and refer to this performance metric as the \emph{frequency of mistakes} for a given concept $c$ in model $m$, $\text{FOM}_m(c)$.

Formally,
\begin{equation}
    \text{FOM}_m(c) = \frac{1}{N} \sum_{i=1}^{N} \mathds{1} \left[ S_i \notin D_c \wedge L_m(S_i) = c \right],
\end{equation}
where $N=10,948$ is the total number of tests (prompts) conducted on $L$ during the drawing selection phase on each modality.

\begin{figure*}[htbp]
	\centering
        \begin{minipage}{0.45\textwidth}
		\centering
        \includegraphics[width=0.80\linewidth]{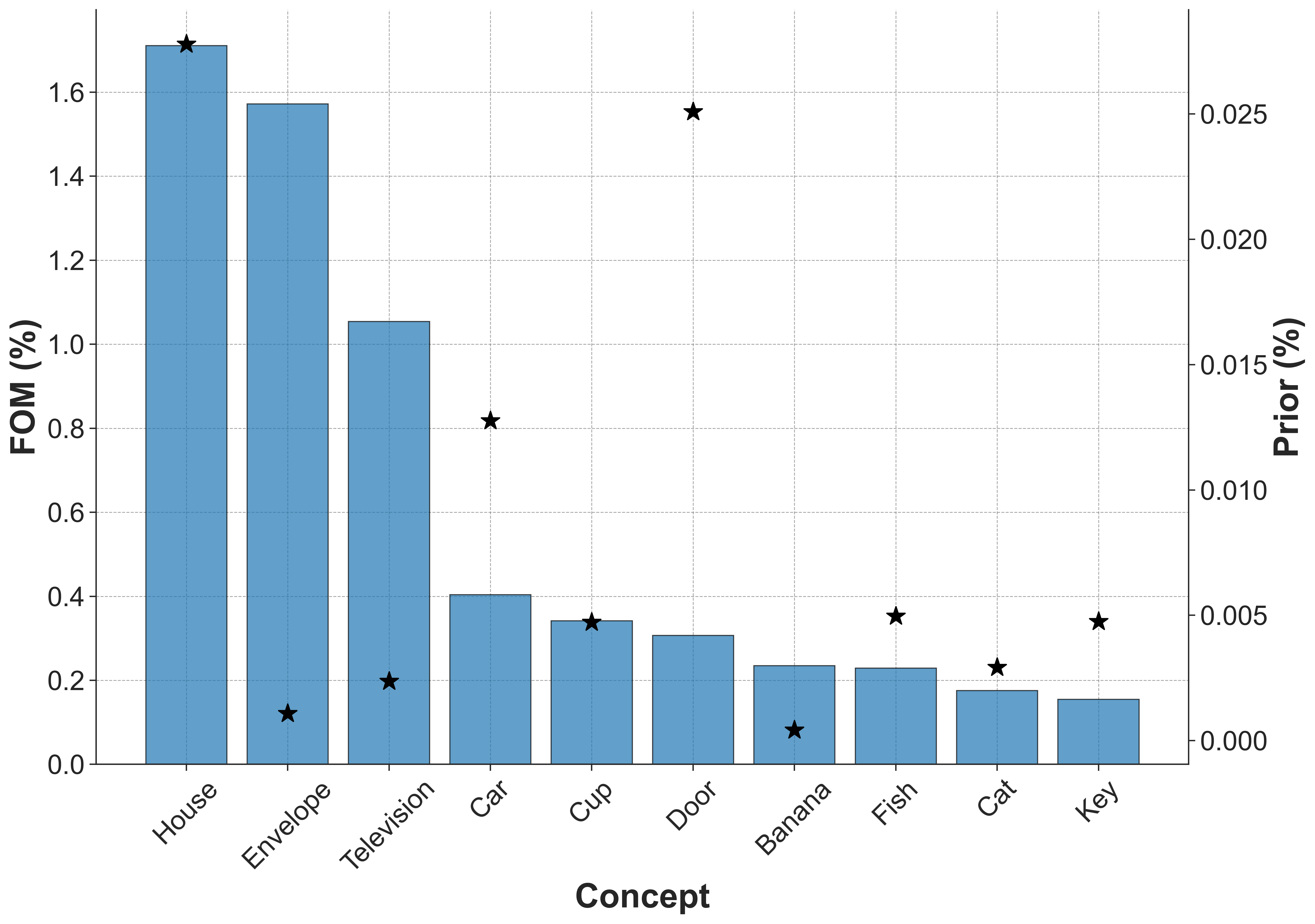}
	\end{minipage}
       $\:\:\:\:\:\:\:\:$
	\begin{minipage}{0.45\textwidth}
		\centering
        \includegraphics[width=0.80\linewidth]{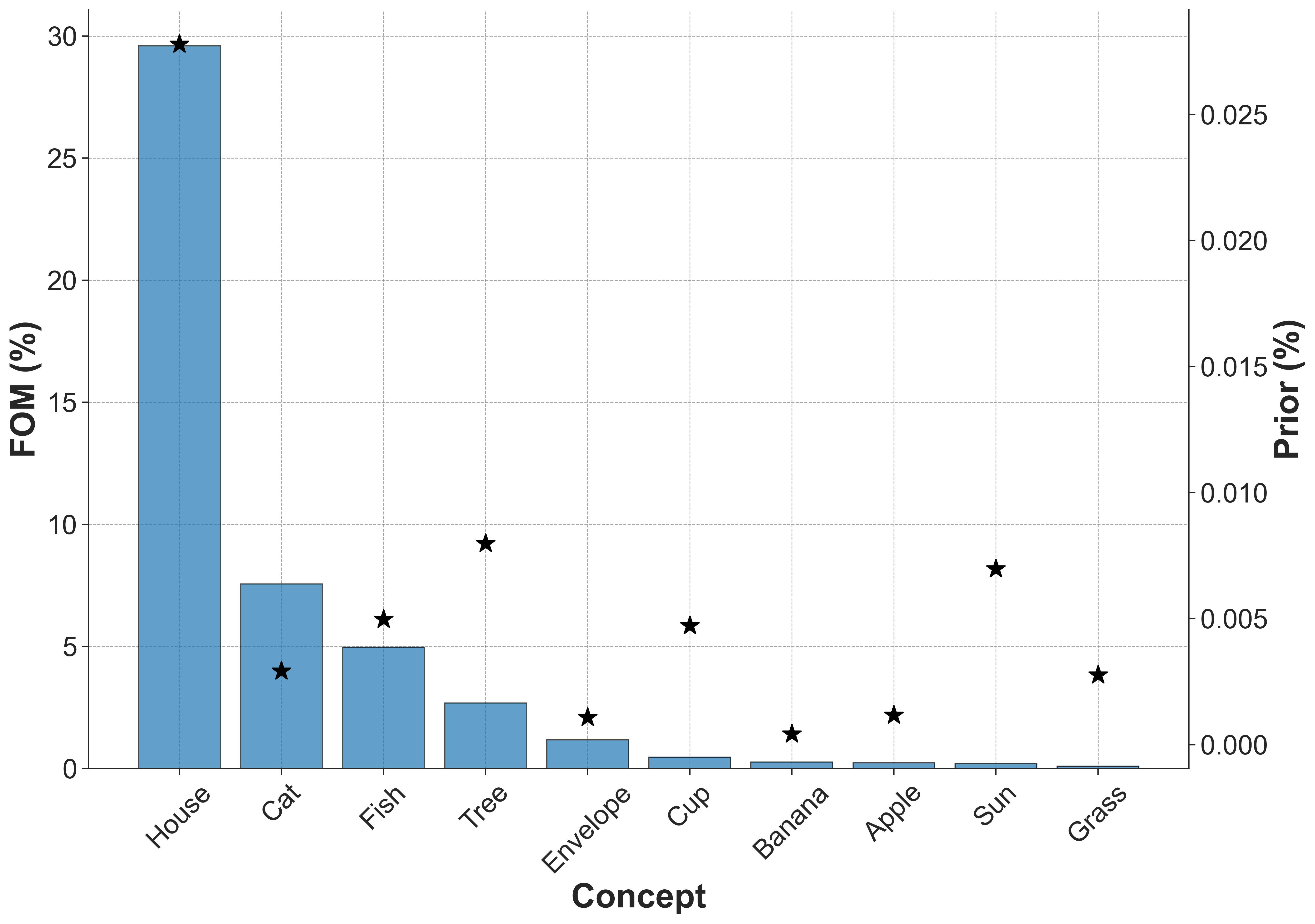} 
	\end{minipage}
	\caption{Top-10 concepts with the highest frequency of mistakes (averaged across the models) in the visual-based modality (images) (left) and text-based modality (coordinates) (right). The little star represents the prior probability for each concept.}
	\label{fig:mistakes-both-modalities-side-by-side}
\end{figure*}

To simplify the interpretation of the frequency of mistakes for a given concept, we average the $\text{FOM}(c)$ across all models. We also explore whether there is a relationship between the frequency of mistakes and the prior probability of each concept. We have included in Tables~8 to~19 of the \suptext{} the confusion matrices for each model and modality. These tables show how well the model performs across various concepts by detailing the true positives and the frequency of errors for each concept. 
Figure~\ref{fig:mistakes-both-modalities-side-by-side} shows that the vision modality exhibits a lower percentage of observed mistakes than the coordinate-based modality. 

Interestingly, as shown in Figure \ref{fig:acc-both-modalities-side-by-side}, the concept \concept{house} in both modality representations, \concept{television} only in the visual-based modality, and \concept{cat} only in the text-based modality, shows the highest accuracy. However, these concepts also have the highest frequency of mistakes, indicating that while they are often correctly identified, they are also frequently guessed when wrong. This indicates that although these concepts are generally easily recognizable, variations in attributes like size and shape may introduce ambiguities that complicate the identification of these concepts. In other words, the models often guess these concepts, whether they are correct or not. 


When calculating the Pearson correlation between the frequency of mistakes and the prior probability, we obtain a correlation of \num{0.914} for the coordinate-based modality and \num{0.434} for the vision-based modality for all concepts. This suggests that in the textual modality, the learner is more susceptible to responding based on their pre-existing biases when confronted with unfamiliar concepts. In contrast, this tendency is reduced in visual representation.

\subsection{Teaching Size}

To calculate the teaching size for each concept, we set $T$ to 1, $\rho$ to \num{0.5}, and $N$ to \num{50}, meaning that a correct identification needs to happen at least 25 times out of 50 trials even with some stochasticity in the model. The aim is to determine the simplest drawing for each modality representation that the learner can identify consistently in at least \num{25} out of \num{50} trials. We highlight that this procedure is different from the one conducted in the previous sections, where the results came from the drawing selection phase.

We present the results for teaching size of images and coordinates in Tables~5 and~6 in the \suptext. Table~7 of the \suptext{} shows the respective simplest drawings identified for each concept, modality and model. The data suggest that, on average, the teaching size values for coordinates (\num{11.46}, SD=\num{8.60}) with successful identification (\num{12}) are higher than those for images (\num{6.73}, SD=\num{2.25}) with successful identification (\num{20}), regardless of the model. Even when considering only the 12 concepts that are well identified using coordinates, the mean teaching size remains lower for images. This indicates that there is no absolute invariance, answering our question Q1 in the negative. In other words, the number of strokes required for a concept to be identified by the learners is generally higher when using textual coordinates compared to bitmap images.

\begin{table*}[htbp]
    \vspace{0.5cm}
	\centering
	\caption{Concept teaching size comparison for images and for coordinates, showing Kendall Rank correlation coefficient for the subset of concepts that are identified (*), and Pearson correlation between the accuracy for all concepts.}\label{tab:TS}
	\begin{tabularx}{\textwidth}{lX X p{0.8cm} p{0.8cm}}
		\toprule
		Model   & Order for Images                                                                                                                                    & Order for Coordinates                                                                                                                                   & Rank* & Pears  \\
		\midrule
		Claude  & \concept{cup} $<$ \concept{house} = \concept{fish} $<$ \concept{envelope} $<$ \concept{apple} = \concept{sun} $<$ \concept{cat} $<$ \concept{grass} & \concept{house} $<$ \concept{envelope} $<$ \concept{apple} $<$ \concept{fish} $<$ \concept{cat} $<$ \concept{sun} $<$ \concept{cup} $<$ \concept{grass} & 0.36 & 0.65                     \\
		Gemini  & \concept{envelope} = \concept{house} = \concept{sun} = \concept{grass} = \concept{fish} $<$ \concept{banana} $<$ \concept{tree} = \concept{cat}     & \concept{envelope} $<$ \concept{house} $<$ \concept{sun} = \concept{tree} $<$ \concept{fish} $<$ \concept{banana} $<$ \concept{grass} $<$ \concept{cat} & 0.57 & 0.21                     \\
		GPT-4o     & \concept{tree} $<$ \concept{house} $<$ \concept{envelope} $<$ \concept{apple} $<$ \concept{cat}                                                     & \concept{envelope} $<$ \concept{house} $<$ \concept{cat} $<$ \concept{tree} $<$ \concept{apple}                                                         & 0.00 & 0.67                     \\
        GPT-4T    & \concept{envelope} $<$ \concept{house} $<$ \concept{fish} $<$ \concept{cat} $<$ \concept{tree} $<$ \concept{car}                                                     & \concept{envelope} $=$ \concept{house} $<$ \concept{fish} $=$ \concept{tree} $<$ \concept{cat} $<$ \concept{car}                                                         & 0.87 & 0.45                      \\
		Llama   & \concept{envelope} = \concept{house} $<$ \concept{cat}                                                                                              & \concept{envelope} = \concept{house} $<$ \concept{cat}                                                                                                  & 1.00 & 0.50                      \\
		Pixtral & \concept{house} $<$ \concept{cat}                                                                                                                   & \concept{house} $<$ \concept{cat}                                                                                                                       & 1.00 & 0.63                     \\
		\bottomrule
	\end{tabularx}
\end{table*}

Furthermore, it is important to highlight a weak, though similar, negative correlation between the teaching size and the prior of each concept across both modalities. The correlation coefficients are \num{-0.021} for coordinates and \num{-0.338} for images, over all concepts and models. This suggests that, in the image modality, the more common a concept is, the simpler its drawings need to be for the learner to consistently identify it.

Interestingly, looking at Table \ref{tab:TS}, the teaching size still ranks concepts in a relatively similar order between images and coordinates, but the strength of this relationship varies across models. The strongest agreement is observed in Llama and Pixtral, both of which exhibit a perfect Kendall rank correlation of 1.0, meaning their rankings are identical across the two modalities. GPT-4 Turbo (shortened as GPT-4T) also exhibits a high correlation (0.87), suggesting strong alignment in concept difficulty ordering between images and coordinates.

However, other models show lower correlations, with Claude at 0.36, Gemini at 0.57, and GPT-4o displaying no correlation (0.0) between the two rankings. To control for the influence of concept priors on teaching size, we performed an ordinary least squares regression of the teaching sizes (for both modalities) on the corresponding concept priors derived from Google Books Ngram frequencies. This yielded residuals representing the portion of teaching size not explained by prior familiarity. We then calculated the Kendall rank correlation between these residuals from different modalities and found similar correlation values.

These results indicate that while some models maintain an invariant notion of teaching size across modalities, others exhibit some discrepancies, although the number of concepts is small. The accuracy correlation between all concepts is a more robust metric, and it also calculates how well concept-wise accuracies align between the two modalities. Claude and GPT-4o exhibit relatively high accuracy correlations (0.65 and 0.67, respectively), suggesting that despite their lower Kendall rank correlations, the overall accuracy patterns remain similar. Meanwhile, Gemini and GPT-4T have lower accuracy correlations (0.21 and 0.45), compensated by the better values for ranking. 

Overall, the correlations are never negative, but less or more positive depending on the model. While Llama, Pixtral, and GPT-4T exhibit strong invariance in teaching size ranking across modalities, others do not. 
In general, however, the answer to question Q2 tends to be positive.

\section{Discussion}

In this study, we examined how multimodal models identify the same concepts in two different modalities: image- and coordinate-based drawings. Our findings show that images are generally more effective than coordinates for identifying concepts. In particular, using images led to the recognition of more concepts than when using coordinates, indicating that images are better suited for teaching concepts to a given learner. This is supported by the higher accuracy and lower frequency of mistakes seen with image-based representations. We also use the number of segments as the teaching size to measure the complexity of a concept. Our analysis indicates that the teaching size is again more beneficial for images than coordinates (clearly answering question Q1 negatively), but ranks concepts in similar ways, regardless of the type of drawing used, even when we account for the learner's priors. While there are differences depending on the model, we tend to see a positive answer to question Q2 more often. This suggests that some concepts are naturally easier or more difficult to teach, no matter how they are represented. 

We believe that our study provides a step towards the investigation of a core question in the field of multimodal \Gls{ai}: Whether language models can interpret structured data (like coordinates) as effectively as images. We saw that models perform better with image-based representations, even for simple concepts. This suggests a limitation in current multimodal models that is important for scientific and practical development. The observed invariance in ranking teaching size across modalities suggests that some concept properties are robust regardless of representation. This may help improve cross-modal transfer learning, where models must generalize concepts between formats.

Our machine teaching framework has several practical implications. First, it improves the design and evaluation of multimodal systems by providing a quantitative, model-agnostic way to measure how \say{costly} it is for any vision-language model to learn a new visual concept in different modalities. Second, our work connects cognitive and computational notions of simplicity by providing empirical evidence that segment count, a classic cognitive cue, remains predictive even in state-of-the-art large language models. This contributes to ongoing discussions in \Gls{ai} about whether these models learn conceptual structures or simply memorize patterns. Finally, the framework can support adaptive teaching tools by identifying the simplest representations for individual learning needs.  Thus, it could be used to develop educational software that teaches geometric concepts or visual reasoning using minimal and optimally chosen examples.

Our analysis has to be seen in the light of some limitations. (a)~The study concentrates on a specific set of concepts, which might affect how well the findings apply to other (potentially more complex) concepts. (b)~Our use of the \Gls{rdp} algorithm for drawing simplification streamlines each stroke but does not totally remove any single stroke from the drawing. This should not be much of a limitation as we focus on the simplest concepts. (c)~A factor that can influence the teaching size of a concept is the curvature of its drawings, i.e., the amount by which it deviates from a straight line. In this work, we have chosen not to focus on this aspect, but this could be of interest for future works.

We show that the simplest concepts usually correspond to those that humans intuitively think of as less complex, and this confirms that the simplest concepts are so across modalities. This supports the hypothesis that the representation of concepts in both modalities is tightly connected in the latent space. However, since we operate under a black-box setting with models like \gls{gpt}-4 and others that do not expose their internal representations, we cannot directly inspect or confirm such latent alignments. Some other methods, especially white-box approaches that have access to weights or gradients, could give a definitive answer to this hypothesis. Still, in cases such as \Gls{gpt}-4 or humans, a black-box approach such as the one presented in this paper is the practical course of action. Thus, our results should be viewed as a hypothesis to explain the invariance across modalities and not as a definitive claim.

The code to reproduce our results is available~\cite{code}.

\bibliography{references}

\newpage
\appendix

\section{Prompts Utilized in this Study}\label{appendix:prompts}

In this work, we use two prompt templates to evaluate the effectiveness of concept identification across multiple models--\Gls{gpt}-4, Llama, Gemini, Pixtral, and Claude--using two different modalities: vision-based and text-based representations.

For the visual modality, the drawings are presented as images generated from the sequence of coordinates. The prompt template for this modality involves showing the model the bitmap image of the drawing and asking it to identify the concept depicted in the image.

For the textual modality, the pen stroke vectors are encoded using the \Tikz{} language. This format allows the representation of drawings as a series of coordinates and commands that describe the strokes. The prompt template for this modality involves presenting the model with these \Tikz-encoded coordinates and asking it to identify the concept represented by the strokes.

Both prompts are designed to elicit open-ended responses from the model, allowing it to consider a wide range of possible concepts, including those not in the predefined \num{20}-concept set. This approach ensures that the model's identification process is not constrained by a limited set of options, thereby providing a more comprehensive evaluation of its capabilities in both modalities.

\begin{prompt}[prompt:visual]{Prompt template for the vision-based modality.}
{\small\texttt{
Your task is to identify a concept drawn by hand. You will be provided with an image corresponding to a concept drawn by hand. Your task is to identify, based on the provided picture, the concept that someone has attempted to draw. Please reply only with the name of the concept.
}}\\\\
\textbf{Image URL:} base 64 encoded drawing ($256\times256$)
\end{prompt}

\begin{prompt}[prompt:text]{Prompt template for the text-based modality.}
{\small\texttt{
Your task is to identify a concept drawn by hand. You will be provided a \Tikz picture format corresponding to a concept, where each stroke is indicated by the command 'draw' followed by a series of points in '(x,y)' format. \\
The points are connected by straight lines, denoted by '--'. The strokes collectively represent a concept. Below is the \Tikz picture code enclosed within triple backticks: \\
'''\{\texttt{\textbf{\Tikz{} code}}\}'''.\\
Your task is to identify, based on the provided \Tikz picture, the concept that someone has attempted to draw. Please reply only with the name of the concept.
}}
\end{prompt}

\paragraph{Example of image for the concept \concept{cat}}

The vision-based modality, on the other hand, involves using images created from the sequence of coordinates from the \dataset dataset. These images are produced by plotting the coordinates with a function that defines the image size as $256\times256$ pixels. The image is then stored in PNG format.

The following is an example of an image representing the  \concept{cat}, extracted from the \dataset{} dataset.

\begin{center}
    \includegraphics[width=0.25\linewidth]{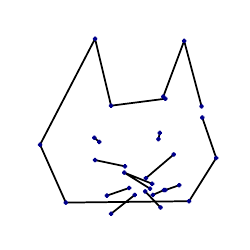}
\end{center}

\paragraph{Example of \Tikz{} code for the concept \concept{cat}}

\Tikz{} is a \LaTeX{} package used for creating graphics programmatically. Because of its way of representing drawings through coordinate-based commands, we used \Tikz{} in the text-modality tests.

Each drawing in the \dataset{} dataset is stored as vectors of distinct pen strokes, represented by sequences of $(x, y)$ coordinates. For each stroke in the drawing, the sequence of points is translated into a \verb|\draw| command. The points are connected using the \texttt{--} operator, which denotes a straight line between two points. Each drawing consists of multiple strokes, and each segment is represented by a separate \verb|\draw| command in \Tikz{}.

The following is an example of the \Tikz{} code of the concept \concept{cat}, extracted from the \dataset{} dataset.

\begin{lstlisting}
\draw (181, 30) -- (121, 12) -- (14, 95) -- (0, 161) -- (42, 255) -- 
    (73, 213) -- (136, 226) -- (236, 194) -- (242, 230) -- (255, 156) -- 
    (218, 38) -- (161, 2) -- (141, 15);
\draw (118, 92) -- (76, 118);
\draw (119, 81) -- (87, 76);
\draw (112, 70) -- (102, 57);
\draw (146, 98) -- (192, 107);
\draw (151, 76) -- (203, 86);
\draw (154, 53) -- (175, 51);
\draw (135, 138) -- (137, 71) -- (123, 81);
\end{lstlisting}

\newpage
\section{Accepted Hyponyms for each Concept} \label{appendix:correspondence}

\begin{center}
	\captionof{table}{Accepted hyponyms for each concept. In this study, we establish a set of accepted hyponyms for each concept. A hyponym is a more specific term within a broader category, and for our purposes, identifying a hyponym is considered correct if it falls under the general expected concept. For instance, if the expected concept is \concept{car}, identifying \concept{ambulance} is still correct because it is a specific type of car. This table lists each concept and its accepted hyponyms. These hyponyms are identified in the drawing selection phase and validated by human inspection.}\label{tab:correspondence}
    \vspace{0.5cm}
	\tablefirsthead{
		\toprule
		Concept ($c$) &                                            Hyponyms ($h(c)$) \\
		\midrule
	}
	\tablehead{
		\toprule
		Concept ($c$) &                                            Hyponyms ($h(c)$) \\
		\midrule
	}
	\tabletail{
		\midrule
		\multicolumn{2}{r}{{Continued on next page.}} \\
	}
	\tablelasttail{
		\bottomrule
	}
	\begin{supertabular}{p{0.3\linewidth} p{0.6\linewidth}}
				
		Apple &                                       Apple logo \\
		\hline
		\multirow{3}{*}{Banana}  &                    Banana peel \\ & Banana pepper \\  & Banana/crescent moon \\
		\hline
		\multirow{5}{*}{Car} &                                          Ambulance \\
		&                                           Truck \\
		&                                      Pickup truck \\
		&                                         Tractor \\
		&                                       Tank \\	
		\hline
		\multirow{8}{*}{Cat} &                                   Cat whiskers \\ &
		Cat face \\ &
		Cat head \\ &
		Cat playing with a ball of yarn \\ &
		Cat playing with a toy \\ &
		Cat/fox \\ &
		House with a cat \\ &
		A cat chasing a mouse \\
		\hline
		\multirow{2}{*}{Computer} &                                           Laptop \\
		&                                   Desktop computer \\
		\hline
		\multirow{9}{*}{Cup} &                                  Glass \\ &
		Broken cup \\ &
		Broken glass \\ &
		Coffee cup \\ &
		Coffee mug \\ &          
		Cup and saucer \\ &
		Cup of coffee \\ &
		Cup/glass \\ &
		Glass and napkin \\ &
		Glass of water \\ &
		Jar \\ &
		Jug \\ &
		Mug \\ &
		Pitcher \\ &
		Wine glass \\
		\hline
		\multirow{6}{*}{Door} &                                           Car door \\ &
		Door with a doorknob \\ &
		Doorway \\ &
		Door with a handle \\ &
		Door ajar \\ &
		Swinging door \\ 
		\hline
		Envelope   &                                          (no hyponyms from the completions)  \\
		\hline
		Fish &                                              Whale \\
		\hline
		Grass &                                              Grass/sawtooth wave \\
		\hline
		Hockey puck &                                              Hockey puck and stick \\        
		\hline
		House                        &                                     Triangle and house \\
		Key &                             Key and knife\\
		\hline
        & \\
		\multirow{3}{*}{Radio}                        &                                         Radio controller \\ & 
		Radio controlled car \\ & 
		Radio cassette player \\
		\hline
		String bean   &                                          (no hyponyms from the completions)  \\
		\hline
		\multirow{4}{*}{Sun} &                                                Sunburst \\ & 
		Starburst \\ & 
		Sun rays \\ & 
		Sun/star \\ 
		\hline
		\multirow{3}{*}{Sword} &             Sword in the stone \\ &
		Knife \\ &
		Khukuri \\
		\hline
		\multirow{8}{*}{Television} &  TV \\ &
		Television/TV/monitor/screen \\ &
		Television/TV/monitor \\ &
		Line graph on a TV screen \\ &
		Computer monitor \\ &
		Monitor \\ &
		Desktop monitor \\ &
		Computer monitor \\ 
		\hline
		\multirow{3}{*}{Tree} &  Palm tree \\ & 
		Christmas tree \\ & 
		Tree branch \\
		\hline
		The Great Wall of China   &                                          (no hyponyms from the completions) \\
	\end{supertabular}
\end{center}

\newpage
\section{Accuracy for each Concept, Model and Modality} \label{appendix:accuracy}

\begin{center}
	\captionof{table}{We evaluate the accuracy of concept identification in the drawing selection phase. Accuracy is determined by the proportion of correctly identified concepts over the total number of evaluations for each concept. A correct identification includes cases where a hyponym of the expected concept is recognized, as established in our accepted hyponym mappings. This table presents the accuracy scores for each concept and model, based on responses from the learners when presented with drawings in both visual and text-based representations. }\label{tab:accuracy-concept-model} 
    \vspace{0.5cm}
	\tablefirsthead{
		\toprule
		Concept ($c$) &                                            Model ($m$) & Modality & Accuracy (\%) \\
		\midrule
	}
	\tablehead{
		\toprule
		Concept ($c$) &                                            Model ($m$) & Modality & Accuracy (\%) \\
		\midrule
	}
	\tabletail{
		\midrule
		\multicolumn{4}{r}{{Continued on next page.}} \\
	}
	\tablelasttail{
		\bottomrule
	}
	\begin{supertabular}{p{0.25\linewidth} p{0.25\linewidth} p{0.25\linewidth} p{0.15\linewidth}}

\multirow{12}{*}{The Great Wall Of China} & Claude & Images & 0.00 \\
 & Claude & Coordinates & 0.00 \\
 & GPT-4 Turbo & Images & 0.00 \\
 & GPT-4 Turbo & Coordinates & 0.00 \\
 & GPT-4o & Images & 0.18 \\
 & GPT-4o & Coordinates & 0.00 \\
 & Gemini & Images & 0.00 \\
 & Gemini & Coordinates & 0.00 \\
 & Llama & Images & 0.00 \\
 & Llama & Coordinates & 0.00 \\
 & Pixtral & Images & 0.00 \\
 & Pixtral & Coordinates & 0.00 \\
\hline
\multirow{12}{*}{String Bean} & Claude & Images & 0.00 \\
& Claude & Coordinates & 0.00 \\
& GPT-4 Turbo & Images & 0.00 \\
& GPT-4 Turbo & Coordinates & 0.00 \\
& GPT-4o & Images & 0.00 \\
& GPT-4o & Coordinates & 0.00 \\
& Gemini & Images & 2.61 \\
& Gemini & Coordinates & 0.00 \\
& Llama & Images & 0.00 \\
& Llama & Coordinates & 0.00 \\
& Pixtral & Images & 0.00 \\
& Pixtral & Coordinates & 0.00 \\
\hline
\multirow{12}{*}{Hockey Puck} & Claude & Images & 0.00 \\
& Claude & Coordinates & 0.00 \\
& GPT-4 Turbo & Images & 0.00 \\
& GPT-4 Turbo & Coordinates & 0.00 \\
& GPT-4o & Images & 0.00 \\
& GPT-4o & Coordinates & 0.00 \\
& Gemini & Images & 16.14 \\
& Gemini & Coordinates & 0.00 \\
& Llama & Images & 0.37 \\
& Llama & Coordinates & 0.00 \\
& Pixtral & Images & 0.19 \\
& Pixtral & Coordinates & 0.00 \\
\hline
\multirow{12}{*}{Grass} & Claude & Images & 1.27 \\
& Claude & Coordinates & 2.55 \\
& GPT-4 Turbo & Images & 0.00 \\
& GPT-4 Turbo & Coordinates & 0.00 \\
& GPT-4o & Images & 5.10 \\
& GPT-4o & Coordinates & 0.00 \\
& Gemini & Images & 35.35 \\
& Gemini & Coordinates & 1.91 \\
& Llama & Images & 1.91 \\
& Llama & Coordinates & 0.00 \\
& Pixtral & Images & 3.50 \\
& Pixtral & Coordinates & 0.00 \\
\hline
\multirow{12}{*}{Door} & Claude & Images & 21.82 \\
 & Claude & Coordinates & 0.00 \\
 & GPT-4 Turbo & Images & 8.05 \\
 & GPT-4 Turbo & Coordinates & 0.00 \\
 & GPT-4o & Images & 3.38 \\
 & GPT-4o & Coordinates & 0.00 \\
 & Gemini & Images & 18.44 \\
 & Gemini & Coordinates & 0.00 \\
 & Llama & Images & 3.12 \\
 & Llama & Coordinates & 0.00 \\
 & Pixtral & Images & 3.64 \\
 & Pixtral & Coordinates & 0.00 \\
\hline
\multirow{12}{*}{Radio} & Claude & Images & 14.82 \\
& Claude & Coordinates & 0.00 \\
& GPT-4 Turbo & Images & 6.40 \\
& GPT-4 Turbo & Coordinates & 0.00 \\
& GPT-4o & Images & 1.87 \\
& GPT-4o & Coordinates & 0.00 \\
& Gemini & Images & 52.42 \\
& Gemini & Coordinates & 0.00 \\
& Llama & Images & 13.57 \\
& Llama & Coordinates & 0.00 \\
& Pixtral & Images & 3.28 \\
& Pixtral & Coordinates & 0.00 \\
\hline
\multirow{12}{*}{Car} & Claude & Images & 17.60 \\
& Claude & Coordinates & 0.00 \\
& GPT-4 Turbo & Images & 14.29 \\
& GPT-4 Turbo & Coordinates & 3.95 \\
& GPT-4o & Images & 4.46 \\
& GPT-4o & Coordinates & 0.00 \\
& Gemini & Images & 42.47 \\
& Gemini & Coordinates & 0.00 \\
& Llama & Images & 7.14 \\
& Llama & Coordinates & 0.00 \\
& Pixtral & Images & 3.83 \\
& Pixtral & Coordinates & 0.00 \\
\hline
\multirow{12}{*}{Key} & Claude & Images & 16.12 \\
& Claude & Coordinates & 0.00 \\
& GPT-4 Turbo & Images & 20.80 \\
& GPT-4 Turbo & Coordinates & 0.00 \\
& GPT-4o & Images & 4.55 \\
& GPT-4o & Coordinates & 0.00 \\
& Gemini & Images & 31.40 \\
& Gemini & Coordinates & 0.14 \\
& Llama & Images & 11.71 \\
& Llama & Coordinates & 0.00 \\
& Pixtral & Images & 12.26 \\
& Pixtral & Coordinates & 0.00 \\
& & & \\
& & & \\
& & & \\
& & & \\
& & & \\
& & & \\
\multirow{14}{*}{Banana} & Claude & Images & 32.46 \\
& Claude & Coordinates & 0.00 \\
& GPT-4 Turbo & Images & 15.63 \\
& GPT-4 Turbo & Coordinates & 0.00 \\
& GPT-4o & Images & 5.81 \\
& GPT-4o & Coordinates & 0.00 \\
& Gemini & Images & 29.66 \\
& Gemini & Coordinates & 1.80 \\
& Llama & Images & 15.63 \\
& Llama & Coordinates & 0.00 \\
& Pixtral & Images & 20.44 \\
& Pixtral & Coordinates & 0.00 \\
\hline
\multirow{12}{*}{Sword} & Claude & Images & 27.20 \\
& Claude & Coordinates & 0.00 \\
& GPT-4 Turbo & Images & 19.87 \\
& GPT-4 Turbo & Coordinates & 0.00 \\
& GPT-4o & Images & 6.69 \\
& GPT-4o & Coordinates & 0.00 \\
& Gemini & Images & 39.33 \\
& Gemini & Coordinates & 0.00 \\
& Llama & Images & 20.08 \\
& Llama & Coordinates & 0.00 \\
& Pixtral & Images & 13.60 \\
& Pixtral & Coordinates & 0.00 \\
\hline
\multirow{12}{*}{Computer} & Claude & Images & 19.37 \\
& Claude & Coordinates & 0.17 \\
& GPT-4 Turbo & Images & 17.63 \\
& GPT-4 Turbo & Coordinates & 0.00 \\
& GPT-4o & Images & 12.22 \\
& GPT-4o & Coordinates & 0.00 \\
& Gemini & Images & 48.34 \\
& Gemini & Coordinates & 0.00 \\
& Llama & Images & 13.96 \\
& Llama & Coordinates & 0.00 \\
& Pixtral & Images & 16.93 \\
& Pixtral & Coordinates & 0.00 \\
\hline
\multirow{12}{*}{Cup} & Claude & Images & 37.78 \\
& Claude & Coordinates & 2.91 \\
& GPT-4 Turbo & Images & 13.16 \\
& GPT-4 Turbo & Coordinates & 0.00 \\
& GPT-4o & Images & 7.69 \\
& GPT-4o & Coordinates & 0.00 \\
& Gemini & Images & 50.43 \\
& Gemini & Coordinates & 0.00 \\
& Llama & Images & 37.09 \\
& Llama & Coordinates & 0.00 \\
& Pixtral & Images & 15.21 \\
& Pixtral & Coordinates & 0.00 \\
\hline
\multirow{12}{*}{Fish} & Claude & Images & 45.08 \\
& Claude & Coordinates & 18.47 \\
& GPT-4 Turbo & Images & 25.25 \\
& GPT-4 Turbo & Coordinates & 2.54 \\
& GPT-4o & Images & 5.59 \\
& GPT-4o & Coordinates & 0.17 \\
& Gemini & Images & 42.03 \\
& Gemini & Coordinates & 29.15 \\
& Llama & Images & 45.08 \\
& Llama & Coordinates & 0.00 \\
& Pixtral & Images & 28.98 \\
& Pixtral & Coordinates & 0.00 \\
\hline
\multirow{12}{*}{Tree} & Claude & Images & 38.36 \\
& Claude & Coordinates & 0.16 \\
& GPT-4 Turbo & Images & 30.02 \\
& GPT-4 Turbo & Coordinates & 26.00 \\
& GPT-4o & Images & 27.13 \\
& GPT-4o & Coordinates & 4.33 \\
& Gemini & Images & 39.81 \\
& Gemini & Coordinates & 20.71 \\
& Llama & Images & 39.33 \\
& Llama & Coordinates & 0.16 \\
& Pixtral & Images & 19.90 \\
& Pixtral & Coordinates & 4.33 \\
\hline
\multirow{12}{*}{Apple} & Claude & Images & 51.96 \\
& Claude & Coordinates & 9.89 \\
& GPT-4 Turbo & Images & 36.89 \\
& GPT-4 Turbo & Coordinates & 0.00 \\
& GPT-4o & Images & 30.46 \\
& GPT-4o & Coordinates & 1.57 \\
& Gemini & Images & 42.39 \\
& Gemini & Coordinates & 0.00 \\
& Llama & Images & 52.28 \\
& Llama & Coordinates & 0.00 \\
& Pixtral & Images & 40.03 \\
& Pixtral & Coordinates & 0.00 \\
\hline
\multirow{12}{*}{Sun} & Claude & Images & 61.48 \\
& Claude & Coordinates & 15.78 \\
& GPT-4 Turbo & Images & 40.37 \\
& GPT-4 Turbo & Coordinates & 0.00 \\
& GPT-4o & Images & 18.79 \\
& GPT-4o & Coordinates & 0.00 \\
& Gemini & Images & 65.66 \\
& Gemini & Coordinates & 40.60 \\
& Llama & Images & 45.71 \\
& Llama & Coordinates & 0.00 \\
& Pixtral & Images & 27.61 \\
& Pixtral & Coordinates & 0.70 \\
\hline
\multirow{12}{*}{Television} & Claude & Images & 69.49 \\
& Claude & Coordinates & 0.00 \\
& GPT-4 Turbo & Images & 43.86 \\
& GPT-4 Turbo & Coordinates & 0.00 \\
& GPT-4o & Images & 47.29 \\
& GPT-4o & Coordinates & 0.00 \\
& Gemini & Images & 74.37 \\
& Gemini & Coordinates & 0.00 \\
& Llama & Images & 66.61 \\
& Llama & Coordinates & 0.00 \\
& Pixtral & Images & 52.89 \\
& Pixtral & Coordinates & 0.00 \\
& & & \\
& & & \\
\multirow{12}{*}{Envelope} & Claude & Images & 62.01 \\
& Claude & Coordinates & 9.61 \\
& GPT-4 Turbo & Images & 60.48 \\
& GPT-4 Turbo & Coordinates & 5.46 \\
& GPT-4o & Images & 25.76 \\
& GPT-4o & Coordinates & 19.87 \\
& Gemini & Images & 69.87 \\
& Gemini & Coordinates & 0.66 \\
& Llama & Images & 60.92 \\
& Llama & Coordinates & 4.15 \\
& Pixtral & Images & 49.56 \\
& Pixtral & Coordinates & 0.00 \\
\hline
\multirow{12}{*}{Cat} & Claude & Images & 63.05 \\
& Claude & Coordinates & 23.13 \\
& GPT-4 Turbo & Images & 56.42 \\
& GPT-4 Turbo & Coordinates & 2.26 \\
& GPT-4o & Images & 56.84 \\
& GPT-4o & Coordinates & 10.72 \\
& Gemini & Images & 74.61 \\
& Gemini & Coordinates & 0.28 \\
& Llama & Images & 68.27 \\
& Llama & Coordinates & 78.00 \\
& Pixtral & Images & 58.82 \\
& Pixtral & Coordinates & 18.34 \\
\hline
\multirow{12}{*}{House} & Claude & Images & 64.24 \\
& Claude & Coordinates & 36.67 \\
& GPT-4 Turbo & Images & 56.49 \\
& GPT-4 Turbo & Coordinates & 63.78 \\
& GPT-4o & Images & 54.67 \\
& GPT-4o & Coordinates & 31.89 \\
& Gemini & Images & 64.24 \\
& Gemini & Coordinates & 4.33 \\
& Llama & Images & 60.59 \\
& Llama & Coordinates & 32.12 \\
& Pixtral & Images & 66.06 \\
& Pixtral & Coordinates & 39.64 \\
\end{supertabular}
\end{center}

\newpage
\section{Teaching size for each Concept, Model and Modality} \label{appendix:ts}

\begin{center}
	\captionof{table}{This table presents the teaching size (TS) for each concept in the image modality. The teaching size represents the minimal number of segments required in a drawing for a learner (i.e., a large language model) to recognize the concept with a probability of at least $\rho$ over $N$ independent trials. A lower teaching size indicates a simpler representation of the concept that is still consistently identifiable by the model.
}\label{tlb:ts-images}
	\tablefirsthead{ \toprule Concept & Model & $\text{TS}_{0.5,50}(c)$ & Correct \\ \midrule }
	\tablehead{ \toprule Concept & Model & $\text{TS}_{0.5,50}(c)$ & Correct \\ \midrule }
	\tabletail{ \midrule \multicolumn{4}{r}{{Continued on next page.}} \\ } \tablelasttail{ \bottomrule }
	\begin{supertabular}
    {p{0.25\linewidth} p{0.2\linewidth} p{0.25\linewidth} p{0.2\linewidth}}

\multirow{6}{*}{Apple}      & Claude               & 7            & 50           \\
                            & Gemini                    & 9            & 50           \\
                            & GPT-4 Turbo                        & 9            & 50           \\
                            & GPT-4o                        & 7            & 43           \\
                            & Llama & 8            & 40           \\
                            & Pixtral                    & 10           & 31           \\
\hline\multirow{6}{*}{Banana}     & Claude               & 8            & 50           \\
                            & Gemini                    & 8            & 50           \\
                            & GPT-4 Turbo                        & 10            & 50           \\
                            & GPT-4o                        & 15           & 45           \\
                            & Llama & 10           & 32           \\
                            & Pixtral                    & 9            & 32           \\
\hline\multirow{6}{*}{Car}        & Claude               & 14           & 50           \\
                            & Gemini                    & 10           & 50           \\
                            & GPT-4 Turbo                        & 19            & 50           \\
                            & GPT-4o                        & 23           & 50           \\
                            & Llama & 18           & 27           \\
                            & Pixtral                    & 25           & 27           \\
\hline\multirow{6}{*}{Cat}        & Claude               & 9            & 38           \\
                            & Gemini                    & 9            & 50           \\
                            & GPT-4 Turbo                        & 11            & 50           \\
                            & GPT-4o                        & 9            & 48           \\
                            & Llama & 9            & 33           \\
                            & Pixtral                    & 9            & 39           \\
\hline\multirow{6}{*}{Computer}   & Claude               & 11           & 50           \\
                            & Gemini                    & 3            & 42           \\
                            & GPT-4 Turbo                        & 6            & 50           \\
                            & GPT-4o                        & 6            & 48           \\
                            & Llama & 10           & 34           \\
                            & Pixtral                    & 8            & 44           \\
\hline\multirow{6}{*}{Cup}        & Claude               & 4            & 46           \\
                            & Gemini                    & 4            & 27           \\
                            & GPT-4 Turbo                        & 13            & 50           \\
                            & GPT-4o                        & 5            & 50           \\
                            & Llama & 6            & 34           \\
                            & Pixtral                    & 11           & 32           \\
\hline\multirow{6}{*}{Door}       & Claude               & 4            & 37           \\
                            & Gemini                    & 5            & 50           \\
                            & GPT-4 Turbo                        & 7            & 45           \\
                            & GPT-4o                        & 4            & 47           \\
                            & Llama & 16           & 26           \\
                            & Pixtral                    & 10           & 36           \\
\hline\multirow{6}{*}{Envelope}   & Claude               & 6            & 50           \\
                            & Gemini                    & 5            & 34           \\
                            & GPT-4 Turbo                        & 5            & 50           \\
                            & GPT-4o                        & 6            & 50           \\
                            & Llama & 6            & 42           \\
                            & Pixtral                    & 6            & 37           \\
\multirow{6}{*}{Fish}       & Claude               & 5            & 44           \\
                            & Gemini                    & 5            & 50           \\
                            & GPT-4 Turbo                        & 9            & 50           \\
                            & GPT-4o                        & 9            & 46           \\
                            & Llama & 7            & 40           \\
                            & Pixtral                    & 6            & 42           \\
\hline\multirow{5}{*}{Grass}      & Claude               & 14           & 33           \\
                            & Gemini                    & 5            & 39           \\
                            & GPT-4o                        & 11           & 47           \\
                            & Llama & 19           & 36           \\
                            & Pixtral                    & 11           & 28           \\
\hline Hockey Puck                 & Gemini                    & 13           & 50           \\
\hline\multirow{6}{*}{House}      & Claude               & 5            & 46           \\
                            & Gemini                    & 5            & 50           \\
                            & GPT-4 Turbo                        & 6            & 50           \\
                            & GPT-4o                        & 5            & 28           \\
                            & Llama & 6            & 47           \\
                            & Pixtral                    & 6            & 50           \\
\hline\multirow{6}{*}{Key}        & Claude               & 10           & 38           \\
                            & Gemini                    & 10           & 50           \\
                            & GPT-4 Turbo                        & 11            & 50           \\
                            & GPT-4o                        & 11           & 32           \\
                            & Llama & 15           & 30           \\
                            & Pixtral                    & 14           & 46           \\
\hline\multirow{6}{*}{Radio}      & Claude               & 10           & 33           \\
                            & Gemini                    & 5            & 50           \\
                            & GPT-4 Turbo                        & 17            & 50           \\
                            & GPT-4o                        & 16           & 26           \\
                            & Llama & 10           & 29           \\
                            & Pixtral                    & 19           & 36           \\
\hline String Bean                 & Gemini                    & 13           & 49           \\
\hline\multirow{6}{*}{Sun}        & Claude               & 7            & 46           \\
                            & Gemini                    & 5            & 30           \\
                            & GPT-4 Turbo                        & 7            & 50           \\
                            & GPT-4o                        & 9            & 39           \\
                            & Llama & 7            & 25           \\
                            & Pixtral                    & 9            & 45           \\
\hline\multirow{6}{*}{Sword}      & Claude               & 6            & 50           \\
                            & Gemini                    & 5            & 50           \\
                            & GPT-4 Turbo                        & 7            & 50           \\
                            & GPT-4o                        & 8            & 50           \\
                            & Llama & 7            & 39           \\
                            & Pixtral                    & 7            & 46           \\

\hline\multirow{6}{*}{Television} & Claude               & 6            & 49           \\
                            & Gemini                    & 5            & 50           \\
                            & GPT-4 Turbo                        & 7            & 50           \\
                            & GPT-4o                        & 6            & 49           \\
                            & Llama & 6            & 47           \\
                            & Pixtral                    & 6            & 46           \\
                            & & & \\
                            & & & \\
                            & & & \\
                            & & & \\
                            & & & \\
                            & & & \\
\multirow{6}{*}{Tree}       & Claude               & 7            & 41           \\
                            & Gemini                    & 9            & 49           \\
                            & GPT-4 Turbo                        & 14            & 50           \\
                            & GPT-4o                        & 4            & 48           \\
                            & Llama & 9            & 26           \\
                            & Pixtral                    & 10           & 26           \\
\end{supertabular}
\end{center}
\newpage
\begin{center}
	\captionof{table}{This table presents the teaching size (TS) for each concept in the coordinates modality. The teaching size represents the minimal number of segments required in a drawing for a learner (i.e., a large language model) to recognize the concept with a probability of at least $\rho$ over $N$ independent trials. A lower teaching size indicates a simpler representation of the concept that is still consistently identifiable by the model.
}\label{tlb:ts-coordinates}
\vspace{0.5cm}
	\tablefirsthead{ \toprule Concept & Model & $\text{TS}_{0.5,50}(c)$ & Correct \\ \midrule }
	\tablehead{ \toprule Concept & Model & $\text{TS}_{0.5,50}(c)$ & Correct) \\ \midrule }
	\tabletail{ \midrule \multicolumn{4}{r}{{Continued on next page.}} \\ } \tablelasttail{ \bottomrule }
	\begin{supertabular}
    {p{0.25\linewidth} p{0.2\linewidth} p{0.25\linewidth} p{0.2\linewidth}}

\multirow{2}{*}{Apple}    & Claude               & 10           & 31           \\
                          & GPT-4o                        & 33           & 31           \\
\hline Banana                    & Gemini                    & 13           & 48           \\
\hline Car                       & GPT-4 Turbo                              & 31           & 50           \\
\hline\multirow{6}{*}{Cat}      & Claude               & 12           & 41           \\
                          & Gemini                    & 19           & 46           \\
                          & GPT-4 Turbo                              & 20           & 50           \\
                          & GPT-4o                        & 13           & 32           \\
                          & Llama & 7            & 26           \\
                          & Pixtral                     & 21           & 28           \\
\hline Cup                       & Claude               & 15           & 26           \\
\hline \multirow{5}{*}{Envelope} & Claude               & 4            & 29           \\
                          & Gemini                    & 5            & 28           \\
                          & GPT-4 Turbo                              & 5            & 50           \\
                          & GPT-4o                        & 4            & 28           \\
                          & Llama & 4            & 29           \\
\hline \multirow{3}{*}{Fish}     & Claude               & 11           & 36           \\
                          & Gemini                    & 8            & 48           \\
                          & GPT-4 Turbo                              & 15           & 32           \\
\hline \multirow{2}{*}{Grass}    & Claude               & 37           & 27           \\
                          & Gemini                    & 17           & 44           \\
\hline \multirow{6}{*}{House}    & Claude               & 3            & 26           \\
                          & Gemini                    & 6            & 50           \\
                          & GPT-4 Turbo                              & 5           & 50           \\
                          & GPT-4o                        & 5            & 30           \\
                          & Llama & 4            & 41           \\
                          & Pixtral                     & 5            & 43           \\
\hline \multirow{2}{*}{Sun}      & Claude               & 13           & 42           \\
                          & Gemini                    & 7            & 50           \\
\hline \multirow{3}{*}{Tree}     & Gemini                    & 7            & 46           \\
                          & GPT-4 Turbo                              & 15           & 47           \\
                          & GPT-4o                        & 15           & 26   \\

\end{supertabular}
\end{center}

\newpage
\section{Original and Simplest Drawing for each Model and Modality}

\begin{center}
	\captionof{table}{Original and simplest drawing for each concept, modality and model. For each concept, the table includes both the original drawing and its simplified version, as processed by the \glsentrylong{rdp} algorithm. The original drawings are those directly sourced from the \dataset{} dataset. In contrast, the simplest drawings result from iterative simplification, which reduces the number of segments while preserving the essential characteristics of the concept. This simplified version represents the minimal form that each learner (i.e., large language model) can still recognize with a high probability (as per the definition of teaching size of this work). By comparing these drawings, we can better understand the inherent simplicity or complexity of each concept and how it translates across visual and textual representations.}\label{tab:ts-image-concepts}
    \vspace{0.5cm}
	\tablefirsthead{ \toprule Concept & Model & Original (images) & Simplified (images) & & Original (coordinates) & Simplified (coordinates) \\ \midrule }
	\tablehead{ \toprule Concept & Model & Original (images) & Simplified (images) & & Original (coordinates) & Simplified (coordinates) \\ \midrule }
	\tabletail{ \midrule \multicolumn{7}{r}{{Continued on next page.}} \\ } \tablelasttail{ \bottomrule }
\begin{supertabular}
		{p{0.1\linewidth} p{0.1\linewidth} p{0.15\linewidth} p{0.15\linewidth} p{0.01\linewidth} p{0.15\linewidth} p{0.15\linewidth}}

		\multirow{25}{*}{Apple} & Claude &
		\includegraphics[width=\linewidth]{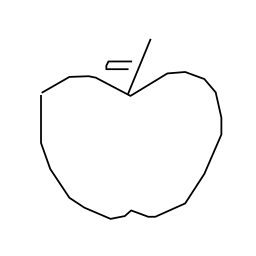}
		&
		\includegraphics[width=\linewidth]{
			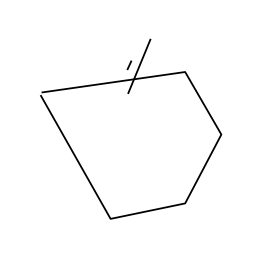
		}
		& &
		\includegraphics[width=\linewidth]{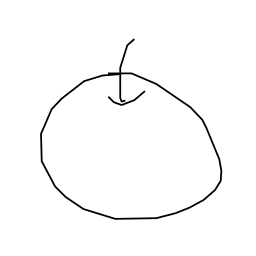}
		&
		\includegraphics[width=\linewidth]{
			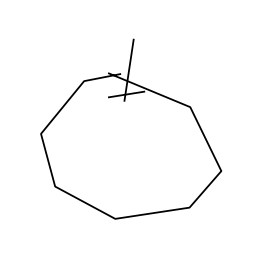
		}
		\\

		& Gemini &
		\includegraphics[width=\linewidth]{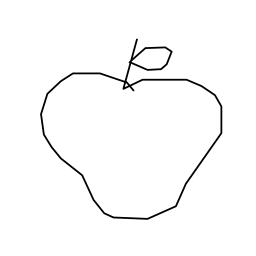}
		&
		\includegraphics[width=\linewidth]{
			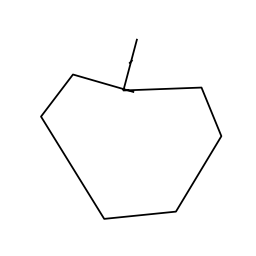
		}
		& & & \\

		& GPT-4 Turbo &
		\includegraphics[width=\linewidth]{images/drawings/5281867072798720_e_2.png}
		&
		\includegraphics[width=\linewidth]{
			images/drawings/5281867072798720_e_39.png
		}
		& & & \\

		& GPT-4o &
		\includegraphics[width=\linewidth]{images/drawings/4503925833072640_e_2.png}
		&
		\includegraphics[width=\linewidth]{
			images/drawings/4503925833072640_e_40.png
		}
		& &
		\includegraphics[width=\linewidth]{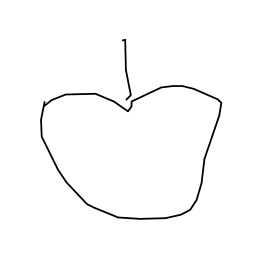}
		&
		\includegraphics[width=\linewidth]{images/drawings/6198979790372864_e_2.png}
		\\

		& Llama &
		\includegraphics[width=\linewidth]{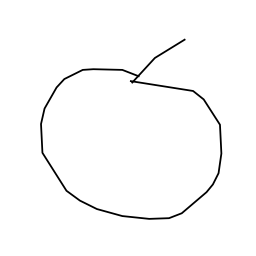}
		&
		\includegraphics[width=\linewidth]{
			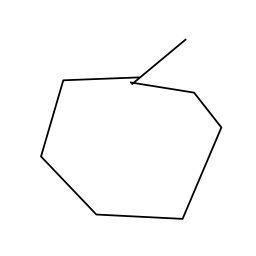
		}
		& & & \\

		& Pixtral &
		\includegraphics[width=\linewidth]{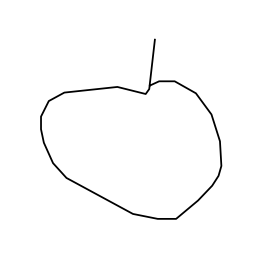}
		&
		\includegraphics[width=\linewidth]{
			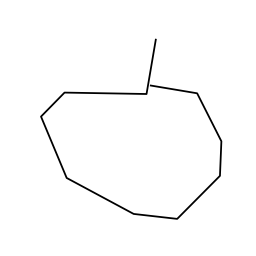
		}
		& & & \\ \multirow{25}{*}{Banana} & Claude &
		\includegraphics[width=\linewidth]{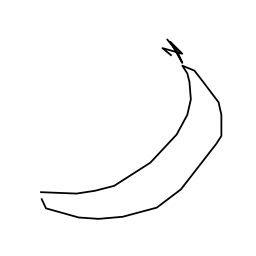}
		&
		\includegraphics[width=\linewidth]{
			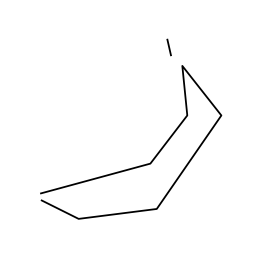
		}
		& & & \\
\end{supertabular}
\end{center}

\begin{landscape}

\section{Confusion Tables of the Classification}

\begin{table}[h]
    \centering
    \caption{Confusion matrix for the Claude model, showing the number of times each concept is accurately predicted or misclassified in the visual-based modality (images). Each cell in the matrix represents the count of instances for a specific actual concept versus a predicted concept. The ‘‘Other’’ column shows the number of predictions that do not match any predefined concepts, since the model is allowed to provide open-ended answers.}
    \label{tab:confusion-images-claude}
\resizebox{\linewidth}{!}{%
 & 385 \\
\hline Total & 220 & 300 & 0 & 268 & 0 & 419 & 159 & 244 & 696 & 148 & 0 & 4 & 452 & 507 & 338 & 116 & 97 & 121 & 234 & 168 & 6457 & 10948 \\
\bottomrule
\end{tabular}
}

\end{table}
\begin{table}[ht]
    \centering
    \caption{Confusion matrix for the Claude model, showing the number of times each concept is accurately predicted or misclassified in the text-based modality (coordinates). Each cell in the matrix represents the count of instances for a specific actual concept versus a predicted concept. The ‘‘Other’’ column shows the number of predictions that do not match any predefined concepts, since the model is allowed to provide open-ended answers.}
    \label{tab:confusion-coordinates-claude}
\resizebox{\linewidth}{!}{%
 & 385 \\
\hline Total & 0 & 2048 & 0 & 69 & 0 & 204 & 0 & 35 & 0 & 6 & 0 & 21 & 223 & 1805 & 134 & 1 & 0 & 0 & 99 & 1 & 6302 & 10948 \\
\bottomrule
\end{tabular}
}

\end{table}

\begin{table}[ht]
    \centering
    \caption{Confusion matrix for the Gemini model, showing the number of times each concept is accurately predicted or misclassified in the visual-based modality (images). Each cell in the matrix represents the count of instances for a specific actual concept versus a predicted concept. The ‘‘Other’’ column shows the number of predictions that do not match any predefined concepts, since the model is allowed to provide open-ended answers.}
    \label{tab:confusion-images-gemini}
\resizebox{\linewidth}{!}{%
 & 385 \\
\hline Total & 201 & 277 & 11 & 306 & 0 & 388 & 219 & 249 & 586 & 338 & 88 & 134 & 536 & 346 & 270 & 346 & 343 & 228 & 313 & 75 & 5694 & 10948 \\
\bottomrule
\end{tabular}
}

\end{table}
\begin{table}[ht]
    \centering
    \caption{Confusion matrix for the Gemini model, showing the number of times each concept is accurately predicted or misclassified in the text-based modality (coordinates). Each cell in the matrix represents the count of instances for a specific actual concept versus a predicted concept. The ‘‘Other’’ column shows the number of predictions that do not match any predefined concepts, since the model is allowed to provide open-ended answers.}
    \label{tab:confusion-coordinates-gemini}
\resizebox{\linewidth}{!}{%
 & 385 \\
\hline Total & 38 & 1930 & 0 & 309 & 0 & 36 & 1 & 438 & 0 & 14 & 0 & 12 & 2 & 330 & 2 & 0 & 0 & 1 & 1 & 0 & 7834 & 10948 \\
\bottomrule
\end{tabular}
}

\end{table}

\begin{table}[ht]
    \centering
    \caption{Confusion matrix showing for the GPT-4 Turbo the number of times each concept is accurately predicted or misclassified in the visual-based modality (images). Each cell in the matrix represents the count of instances for a specific actual concept versus a predicted concept. The ‘‘Other’’ column shows the number of predictions that do not match any predefined concepts, since the model is allowed to provide open-ended answers.}
    \label{tab:confusion-images-gpt4-turbo}
\resizebox{\linewidth}{!}{%
 &    385 \\
\midrule
Total &                                                 \num{93} &                                                \num{165} &                                                \num{0} &                                                \num{187} &                                                \num{0} &                                                \num{437} &                                                 \num{99} &                                                \num{190} &                                                \num{270} &                                                \num{154} &                                                \num{0} &                                                \num{0} &                                                \num{402} &                                                \num{307} &                                                \num{235} &                                                \num{105} &                                                 \num{42} &                                                \num{153} &                                                \num{113} &                                                 \num{59} &                                             \num{7937} &  \num{10948} \\
\bottomrule
\end{tabular}
}
\end{table}
\noindent
\begin{table}[ht]
    \centering
    \caption{Confusion matrix for the GPT-4 Turbo model, showing the number of times each concept is accurately predicted or misclassified in the text-based modality (coordinates). Each cell in the matrix represents the count of instances for a specific actual concept versus a predicted concept. The ‘‘Other’’ column shows the number of predictions that do not match any predefined concepts, since the model is allowed to provide open-ended answers.}
    \label{tab:confusion-coordinates-gpt4-turbo}
\resizebox{\linewidth}{!}{%
 &    385 \\
\midrule
Total            &                                                \num{0} &                                                \num{157} &                                                \num{0} &                                                  \num{0} &                                                \num{0} &                                                \num{130} &                                                \num{0} &                                                \num{563} &                                                \num{0} &                                                \num{347} &                                                \num{0} &                                                \num{0} &                                                 \num{19} &                                               \num{4672} &                                                  \num{3} &                                                \num{0} &                                                \num{0} &                                                  \num{3} &                                                \num{0} &                                                \num{0} &                                             \num{5054} &  \num{10948} \\
\bottomrule
\end{tabular}
}

\end{table}

\begin{table}[ht]
    \centering
    \caption{Confusion matrix for the GPT-4o model, showing the number of times each concept is accurately predicted or misclassified in the visual-based modality (images). Each cell in the matrix represents the count of instances for a specific actual concept versus a predicted concept. The ‘‘Other’’ column shows the number of predictions that do not match any predefined concepts, since the model is allowed to provide open-ended answers.}
    \label{tab:confusion-images-gpt4o}
\resizebox{\linewidth}{!}{%
 & 385 \\
\hline Total & 37 & 33 & 0 & 81 & 1 & 282 & 32 & 200 & 378 & 36 & 0 & 16 & 482 & 605 & 194 & 97 & 12 & 44 & 49 & 26 & 8343 & 10948 \\
\bottomrule
\end{tabular}
}

\end{table}
\begin{table}[ht]
    \centering
    \caption{Confusion matrix for the GPT-4o model, showing the number of times each concept is accurately predicted or misclassified in the text-based modality (coordinates). Each cell in the matrix represents the count of instances for a specific actual concept versus a predicted concept. The ‘‘Other’’ column shows the number of predictions that do not match any predefined concepts, since the model is allowed to provide open-ended answers.}
    \label{tab:confusion-coordinates-gpt4o}
\resizebox{\linewidth}{!}{%
 & 385 \\
\hline Total & 0 & 39 & 0 & 0 & 0 & 391 & 0 & 141 & 0 & 19 & 0 & 0 & 300 & 2494 & 32 & 0 & 0 & 0 & 3 & 0 & 7526 & 10945 \\
\bottomrule
\end{tabular}
}

\end{table}

\begin{table}[ht]
    \centering
    \caption{Confusion matrix for the Llama model, showing the number of times each concept is accurately predicted or misclassified in the visual-based modality (images). Each cell in the matrix represents the count of instances for a specific actual concept versus a predicted concept. The ‘‘Other’’ column shows the number of predictions that do not match any predefined concepts, since the model is allowed to provide open-ended answers.}
    \label{tab:confusion-images-llama}
\resizebox{\linewidth}{!}{%
 & 385 \\
\hline Total & 87 & 294 & 0 & 204 & 0 & 509 & 121 & 249 & 582 & 74 & 2 & 6 & 514 & 521 & 340 & 94 & 87 & 85 & 230 & 29 & 6920 & 10948 \\
\bottomrule
\end{tabular}
}

\end{table}
\begin{table}[ht]
    \centering
    \caption{Confusion matrix for the Llama model, showing the number of times each concept is accurately predicted or misclassified in the text-based modality (coordinates). Each cell in the matrix represents the count of instances for a specific actual concept versus a predicted concept. The ‘‘Other’’ column shows the number of predictions that do not match any predefined concepts, since the model is allowed to provide open-ended answers.}
    \label{tab:confusion-coordinates-llama}
\resizebox{\linewidth}{!}{%
 & 385 \\
\hline Total & 0 & 0 & 0 & 0 & 0 & 115 & 0 & 59 & 0 & 0 & 0 & 0 & 6420 & 2301 & 0 & 0 & 0 & 0 & 0 & 0 & 2053 & 10948 \\
\bottomrule
\end{tabular}
}

\end{table}

\begin{table}[ht]
    \centering
    \caption{Confusion matrix for the Pixtral model, showing the number of times each concept is accurately predicted or misclassified in the visual-based modality (images). Each cell in the matrix represents the count of instances for a specific actual concept versus a predicted concept. The ‘‘Other’’ column shows the number of predictions that do not match any predefined concepts, since the model is allowed to provide open-ended answers.}
    \label{tab:confusion-images-pixtral}
\resizebox{\linewidth}{!}{%
 & 385 \\
\hline Total & 125 & 190 & 0 & 122 & 0 & 292 & 77 & 176 & 417 & 50 & 1 & 12 & 455 & 759 & 255 & 108 & 21 & 115 & 96 & 21 & 7656 & 10948 \\
\bottomrule
\end{tabular}
}

\end{table}
\begin{table}[ht]
    \centering
    \caption{Confusion matrix for the Pixtral model, showing the number of times each concept is accurately predicted or misclassified in the text-based modality (coordinates). Each cell in the matrix represents the count of instances for a specific actual concept versus a predicted concept. The ‘‘Other’’ column shows the number of predictions that do not match any predefined concepts, since the model is allowed to provide open-ended answers.}
    \label{tab:confusion-coordinates-pixtral}
\resizebox{\linewidth}{!}{%
 & 385 \\
\hline Total & 0 & 6 & 0 & 15 & 0 & 0 & 0 & 322 & 0 & 1 & 0 & 0 & 467 & 2966 & 10 & 0 & 0 & 0 & 0 & 0 & 7161 & 10948 \\
\bottomrule
\end{tabular}
}

\end{table}

\end{landscape}

\end{document}